%% file: ITCdesign_v30.tex
\documentclass[10pt,twocolumn,twoside]{IEEEtran}	
\usepackage{cite,graphicx,amsmath,amsfonts,amssymb,psfrag,mathrsfs}
\usepackage[usenames]{color}
\usepackage[nolist]{acronym} 
\usepackage{subfigure}
\usepackage[multiple]{footmisc}
\usepackage{flushend}   

\input{mathdef.tex}

\begin{document}

\input{acronymsAM.tex}  

\title{Model Order Selection Based on Information Theoretic Criteria: Design of the Penalty}
\author{Andrea~Mariani,~\IEEEmembership{Member,~IEEE}, Andrea~Giorgetti,~\IEEEmembership{Senior~Member,~IEEE}, and Marco~Chiani,~\IEEEmembership{Fellow,~IEEE}
\thanks{This work was supported in part by the European Union's Seventh
Framework Programme (FP7/2007-2013) under grant agreement CONCERTO
n. 288502 and in part by the Italian Ministry of Education,
Universities and Research (MIUR) under Research Projects of
Significant National Interest PRIN 2011 GRETA.
}
\thanks{The authors are with the 
CIRI-ICT/DEI, University of Bologna,
        Via Venezia 52, 47521 Cesena, ITALY
        (e-mail: \{a.mariani, andrea.giorgetti, marco.chiani\}@unibo.it).}%
}


\date{\today} \maketitle




\begin{abstract}
Information theoretic criteria (\acsu{ITC}) have been widely adopted in engineering and statistics for selecting, among an ordered set of candidate models, the one that better fits the observed sample data. The selected model minimizes a penalized likelihood metric, where the penalty is determined by the criterion adopted. While rules for choosing a penalty that guarantees a consistent estimate of the model order are known, theoretical tools for its design with finite samples have never been provided in a general setting. In this paper, we study model order selection for finite samples under a design perspective, focusing on the \ac{GIC}, which embraces the most common \ac{ITC}.  
The theory is general, and as case studies we consider: a) the problem of estimating the number of signals embedded in \ac{AWGN} by using multiple sensors; 
 b) model selection for the \ac{GLM}, which includes e.g.~the problem of estimating the number of sinusoids in \ac{AWGN}. The analysis reveals a trade-off between the probabilities of overestimating and underestimating the order of the model. We then propose to design the \ac{GIC} penalty to minimize underestimation while keeping the overestimation probability below a specified level. For the considered problems, this method leads to analytical derivation of the optimal penalty for a given sample size. A performance comparison between the penalty optimized GIC and common AIC and BIC is provided, demonstrating the effectiveness of the proposed design strategy.
\end{abstract}
%
{\IEEEkeywords Akaike information criterion, Bayesian information criterion, general linear model, generalized information criterion, information theoretic criteria, model order selection.}
%
 \acresetall  
\section{Introduction}
\label{sec:introduction}

\IEEEPARstart{M}{odel order selection} problems occurring in engineering and statistics are often solved by means of \ac{ITC} \cite{RaoWu:01,BurAnd:02,StoSel:J04}. The selected model order minimizes a penalized likelihood metric, where the penalty is determined by the criterion adopted. The most commonly used criteria are the \ac{AIC} and the \ac{BIC}, which are the forefathers of the classes of criteria derived from \ac{K-L} information and from Bayesian estimation, respectively \cite{BurAnd:02,ChaGho:11}.\footnote{The expression ``\acl{ITC}'' has been originally used referring to the derivation of \ac{AIC} from the \ac{K-L} information \cite{Aka:74}. In literature, approaches based on Bayesian estimation are commonly numbered among the \ac{ITC} due to their form similar to \ac{AIC} (see \eqref{eq:kestITCb} and \eqref{eq:kestITC}), even though they are not derived from information theoretic arguments.}
Despite {the fact} that \ac{ITC} have been largely studied and adopted, there are relatively few works that address the derivation of \ac{ITC} as a design problem. Most of them study the consistency of model order selection deriving the conditions under which asymptotically, for a large number of observations, the correct model order is chosen \cite{Nis:88,FouSri:77,Boz:87,Gas:13}. However, in practice, finite sample sizes are used, and consistency considerations are not sufficient for controlling the error probabilities. Some works empirically study how to set the penalty in specific selection problems \cite{StoSel:J04,BhaDow:77,Atk:80}. For example, in \cite{BhaDow:77} and \cite{Atk:80} the values to be adopted in autoregressive model selection problems are discussed, while in \cite{Nad:10} a modification of \ac{AIC} has been proposed. Some effort has been placed on non-asymptotic penalties for some specific problems, such as Gaussian model selection \cite{Bir:07}.

In this paper, we study \ac{ITC} under a design perspective, focusing on the study of the \ac{GIC}, which embraces most common criteria such as \ac{AIC} and \ac{BIC}. The \ac{GIC} performance analysis for finite sample sizes reveals a trade-off between the probabilities of overestimating and underestimating the order of the model. Thus, we propose to design the \ac{GIC} penalty to minimize underestimation while keeping the overestimation probability below a specified level. As a practical case study, we focus on the classical problem of estimating the number of signals in Gaussian noise, which arises in many statistical signal processing and wireless communication applications. For example, in the context of cognitive radio, the enumeration of active transmissions is of great interest for increasing the spectrum awareness of the \acl{SU} systems \cite{KanGio:12,ChiWin:10}. The most commonly used approaches for solving this problem are the non-parametric model order estimators proposed in \cite{WaxKai:85}, that received a considerable attention in the past decades \cite{WaxKai:85,ZhaKriBai:86,XuKav:95,FisGroMes:02,ChiWin:10,Nad:10,KriNad:09}. As a second example, we consider model order selection for the \ac{GLM}, which can be used, e.g., for estimating the number of sinusoids in \ac{AWGN}, and for model selection in autoregressive processes \cite{Gra:76,Dju:98,Cho:92}. In both cases it is shown that the performance for high \acp{SNR} is determined by noise distribution. 

The contributions of the paper are as follows.
\begin{itemize}
\item We analyze the probability of correct model selection for the \ac{GIC}. This study gives an insight on the performance of \ac{ITC}, relating underestimation and overestimation events to the penalty adopted. This applies to the whole class of \ac{GIC}, including \ac{AIC} and \ac{BIC}.
\item We propose a design strategy for the \ac{GIC} penalty. This approach minimizes underestimation while keeping the overestimation probability below a specified level.
\item We address the problem of estimating the number of sources in white noise applying the \ac{GIC} design approach proposed. For this case, design is based on a new closed-form approximation of the probability of correct model selection for high \ac{SNR}, based on the statistic of the ratio of the largest eigenvalue to the trace of a white central Wishart random matrix. 
\item We address model selection for the \ac{GLM}. In this case, being an analytical form of the correct selection probability not available, we design the penalty by means of tight performance bounds. As an application example, we focus on the problem of estimating the number of sinusoids in \ac{AWGN}.
\end{itemize}

The paper is organized as follows. Model order selection is introduced in Section~\ref{sec:ITC}. In Section~\ref{sec:perfdesign}, we derive the \ac{GIC} performance and propose a design approach, which is applied to the problem of estimating the number of sources in Section~\ref{sec:WK} and to the \ac{GLM} in Section~\ref{sec:GLM}. Numerical results are presented in Section~\ref{sec:numerical_results}.

Throughout the paper, boldface letters denote matrices and vectors, and $\mathbf{X} \sim \CN{\boldsymbol{0}}{\boldsymbol\Sigma}$ denotes a circularly symmetric complex Gaussian random vector with zero mean and covariance matrix ${\boldsymbol{\Sigma}}$. Also, $X \sim\chi_{m}^2$ is a central chi squared distributed \ac{r.v.} with $m$ \acl{d.o.f.}, $X\sim\mathcal{G}\!\left(\kappa,\theta\right)$ is a gamma distributed \ac{r.v.} with shape parameter $\kappa$ and scale parameter $\theta$, and $X\sim\betadist{a}{b}$ is a beta distributed \ac{r.v.} with parameters $a$ and $b$. We denote the \ac{p.d.f.} and \ac{CDF} of the \ac{r.v.} $X$ with $f_{X}(x)$ and $F_{X}(x)$, respectively. The notation $X\stackrel{d}{\approx}Y$ means that the distribution of the \ac{r.v.} $X$ can be approximated by the distribution of the \ac{r.v.} $Y$. Moreover, $\Id{m}$ represents an identity matrix of order $m$, $\tr{\mathbf{A}}$ is the trace of the matrix $\mathbf{A}$, $\transp{(\cdot)}$ and $\hmt{(\cdot)}$ stand, respectively, for simple and Hermitian transposition.

\section{Information Theoretic Criteria for Model Order Selection}
\label{sec:ITC}

In \cite{Aka:74} Akaike first proposed an information theoretic criterion for statistical model selection based on the observation of $\n$ \ac{i.i.d.} samples of the $\p$ dimensional random vector $\mathbf{X}$, generated by the distribution $f\big(\mathbf{X};\thetaq\big)$, where $\thetaq$ is the vector that contains the unknown parameters of the model. The length of $\thetaq$ increases with the model order $q$. Model order selection consists in identifying the model that better fits data among a set of possible models $\left\{\fcond{\mathbf{X}}{\thetak}\right\}_{k\in\mathcal{K}}$, each one characterized by the model order $k$ and the corresponding parameter vector $\thetak$.\footnote{We refer to the $k$-th model also as the $k$-th hypothesis.} 
Throughout the paper, we assume that the true model is included in the model set considered. The analysis of the case in which the true model is misspecified is out of the scope of the paper. We focus, in particular, on the selection problems in which the hypotheses are nested, which means that the $i$-th hypothesis is always contained in the $j$-th one, with $i<j$. The set of the possible values assumed by $k$ is $\mathcal{K}=\left\{0,1,\dots,\qmax \right\}$, where $\qmax$ is the maximum model order considered.

Denoting by $\xv{i} = \transp{\left(x_{i,1},x_{i,2},\dots,x_{i,\p}\right)}$ the $i$-th sample of $\mathbf{X}$, we build the $\p\times\n$ observation matrix
\begin{align} \label{eq:Y}
\Y = \left(\xv{1}| \xv{2}|\cdots|\xv{\n}\right).
\end{align}
We assume that each sample $\xv{i}$ is composed by a signal part $\sv{i}$ corrupted by an additive noise component $\nv{i}$, i.e., $\xv{i}=\sv{i}+\nv{i}$, and we define the \ac{SNR} as $\SNR=\EX{\hmt{\sv{i}}\sv{i}}/\EX{\hmt{\nv{i}}\nv{i}}$, which is assumed to be independent of $i$. According to the general formulation of \ac{ITC}, the model that better fits data is the one that minimizes the metric
\begin{align} \label{eq:kestITCb}
\ITCk &= -2\sum_{i=1}^{\n}\lnfcond{\xv{i}}{\thetakest} + \Pkk{k}
\end{align}
where $\thetakest$ is the \ac{ML} estimate of the vector $\thetak$, and $\Pkk{k}$ is the penalty.\footnote{Using the notation $\Pkk{k}$ we emphasize that the penalty depends on $k$, which is important for the minimization in \eqref{eq:kestITC}. Note that, in general, $\Pkk{k}$ could also depend on other parameters.} Thus, the model order selected is 
\begin{align} \label{eq:kestITC}
\qest &= \arg \min_{k} \ITCk.
\end{align}
Each criterion is defined by its particular penalty which impacts the performance and the complexity of model order selection.

Note that the formulation of the selection problem as in \eqref{eq:kestITCb} and \eqref{eq:kestITC} supports the interpretation of \ac{ITC} as extensions of the \ac{ML} principle in the form of penalized likelihood. In fact, the \ac{ML} approach performs poorly in model order selection problems, always leading to the choice with 
maximum number of unknown parameters \cite{Sch:78}. The penalty is introduced in \eqref{eq:kestITCb} as a cost to account for the increased complexity of the model, related to the presence of unknown parameters that must be estimated \cite{Aka:74,HanYu:01}. Thus, model selection based on \ac{ITC} extends the \ac{ML} approach, in that it takes into account both the estimation (of the unknown parameters) and the decision (among the possible models).

\subsection{Review of fundamental criteria}
\label{sec:ITCreview}

Akaike proposed to select the model which minimizes the \ac{K-L} {divergence} from $\fcond{\Xv}{\thetak}$ to  $\fcond{\Xv}{\thetaq}$. In fact, since
\begin{align}
\q = \arg \min_{k} \, \int \fcond{\Xv}{\thetaq} \ln{\frac{\fcond{\Xv}{\thetaq}}{\fcond{\Xv}{\thetak}}} \,d\Xv
\end{align}
the correct order is the one minimizing the cross entropy 
\begin{align}
-\int \fcond{\Xv}{\thetaq} \lnfcond{\Xv}{\thetak} d\Xv
\end{align}
for which an estimate, under the $k$-th hypothesis, is given by the average log-likelihood with \ac{ML} estimate of the parameters
\begin{align}
-\frac{1}{\n} \sum_{i=1}^{\n} \lnfcond{\xvi}{\thetakest}.
\end{align}

Akaike noted that the average log-likelihood is a biased estimate of the cross entropy, and added a penalty that asymptotically, for large $\n$, compensates the estimation error.
Exploiting the asymptotical chi squared distribution of the log-likelihood, he derived what is now called the \ac{AIC}, that corresponds to \eqref{eq:kestITCb} and \eqref{eq:kestITC} with penalty
\begin{align} \label{eq:PkAIC}
\PAICk = 2\,\phik
\end{align}
where $\phik$ is the number of free parameters in $\thetak$. Thus, the \ac{AIC} metric aims to minimize an unbiased estimate of the \ac{K-L} divergence. However, in many situations it tends to overestimate the order of the model, even asymptotically \cite{Boz:87, BurAnd:02, LiaReg:01, Nad:10, WaxKai:85, XuKav:95, ZhaWon:89, StoSel:J04, Nis:88, Shi:76}.

Alternative \ac{ITC} are derived adopting the Bayesian approach, which chooses the model maximizing the a posteriori probability $\Prob{\left. \thetak \right|\xv{1},\xv{2},\dots,\xv{\n}}$. In this context, the most simple criterion is the \ac{BIC} with penalty\footnote{The \ac{BIC} was originally derived in \cite{Sch:78}, assuming that the observations come from an exponential family distribution.} \cite{Sch:78}
\begin{align} \label{eq:PkBIC}
\PBICk = \phik \,\ln \n.
\end{align}
For large enough samples, \ac{BIC} coincides with the \acs{MDL} criterion, which attempts to construct a model that permits the shortest description of the data \cite{Ris:04}. It has been demonstrated that in some cases \ac{BIC} provides a consistent estimate of the model order \cite{ZhaKriBai:86b,Nis:88,Cas:09}.\footnote{See also Section~\ref{sec:design}}

More generally, a large number of \ac{ITC}, including \ac{AIC} and \ac{BIC}, present a penalty in the form
\begin{align} \label{eq:PkGIC}
\PGICk = \phik \cdot \nu
\end{align}
where $\nu$ can be a constant (as in \eqref{eq:PkAIC}) or a function of other parameters (as in \eqref{eq:PkBIC}). We refer to this criterion as the \ac{GIC} \cite{Nis:84,Nis:88,StoSel:J04,ZhaLiTsa:10}.
It has been shown that consistency of \ac{GIC} can be reached by properly adjusting the parameter $\nu$\cite{Aka:78, WaxKai:85, Nis:88, ZhaKriBai:86}. In particular, it can be demonstrated that it is required, for $\n$ that goes to infinity, that $\nu/\n\rightarrow 0$ to avoid underestimation and $\nu/\ln\ln\n\rightarrow +\infty$ to avoid overestimation \cite{Nis:88}. Further rules can be derived in some specific selection problems \cite{ZhaKriBai:86}. Based on these general results, different criteria have been proposed, such as in \cite{Boz:87}, where $\nu=1+\ln\n$ is used, and in \cite{GioChi:13}, where $\nu=2\,\ln \n$ has been adopted. A summary of the main \ac{ITC} proposed in literature can be found in \cite[Section 4]{RaoWu:01}.

In the next section, we discuss the performance and the design of \ac{GIC} for finite samples, proposing a method for setting $\nu$ given a target maximum probability of overestimation. 

\section{GIC performance and design}
\label{sec:perfdesign}

\subsection{Model selection performance}
\label{sec:perf}

The performance of model order selection is evaluated in terms of probability to correctly detect $\q$, $\Pc\triangleq \Prob{\qest=\q}$, that can be expressed as
\begin{align} \label{eq:Pk}
\Pc=\Pc\!\left(\q,\SNR,\nu\right)=1-\Pover-\Punder
\end{align}
where $\Pover\triangleq\Prob{\qest>\q}$, with $\q\in\{0,1,\dots,\p-2\}$, and $\Punder\triangleq\Prob{\qest<\q}$, with $\q\in\{1,2,\dots,\p-1\}$, are the probabilities of overestimation and underestimation, respectively. Given \eqref{eq:kestITC}, $\Pover$ and $\Punder$ can be expressed as\footnote{These expressions of $\Pover$ and $\Punder$ are based on the fact that in most of model order selection problems $\ITCkk{k}$ is a concave function with a minimum that in case of correct selection corresponds to $k=\q$. This occurs, for example, in the case studies considered in the paper \cite{XuKav:95,ZhaWon:89}.}
\begin{align} \label{eq:Pover_general}
\Pover &\simeq \Prob{ \bigcup_{i=1}^{\qmax-\q} \!\!\big\{\ITCkk{\q+i}<\ITCkk{q}\big\} }\\
\Punder &\simeq \Prob{\, \bigcup_{i=1}^{\q} \big\{\ITCkk{\q-i}<\ITCkk{q}\big\} }.
\label{eq:Punder_general}
\end{align}

Considering $\Pover$, simple upper and lower bounds, $\PoverUB$ and $\PoverLB$, are respectively given by
\begin{align} \label{PoverUB}
\Pover &\leq \sum_{i=1}^{\qmax-q} \mathbb{P}\Big(\ITCkk{q+i}<\ITCkk{q}\!\Big) \nonumber\\
&\approx \sum_{i=1}^{\imax} \mathbb{P}\Big( \ITCkk{q+i}<\ITCkk{q}\! \Big) = \PoverUB \\
\Pover &\geq \!\max_{i\in\{1,\dots,\qmax-q\}} \,\mathbb{P}\Big(\ITCkk{q+i}<\ITCkk{q} \!\Big) \nonumber\\
&\geq \mathbb{P}\Big(\ITCkk{q+1}<\ITCkk{q} \!\Big) = \PoverLB 
\label{PoverLB}
\end{align}
where the sum in \eqref{PoverUB} is truncated to the integer value $\imax$, with $1\leq \imax \leq \qmax -\q$.
The expressions of the bounds in \eqref{PoverUB} and \eqref{PoverLB} are based on the assumption that $\mathbb{P}\big(\ITCkk{q+i}<\ITCkk{q}\big)$ is decreasing with $i$, which is common for \ac{ITC} based model order selection, as in the case studies discussed in the following sections.\footnote{In general, different problems require a different $\imax$. For example, in the Section~\ref{sec:numerical_results} we show that for the problem of estimating the number of signals $\imax=1$ is sufficient for approximating $\Pover$, while for the problem of estimating the number of sinusoids at least $\imax=2$ is required.} Similar considerations can be applied to the analysis of $\Punder$.

When the \ac{SNR} increases it has been noted that $\Punder$ goes to zero, while $\Pover$ converges to a non zero value \cite{ZhaWon:89,Kav:87,FisGroMes:02,Dju:96,Dju:98}.\footnote{In Fig.~\ref{fig:pfapmdGIC} and Fig.~\ref{fig:pfapmdGLM} we show some simulation results that confirm this effect.} This means that in the high \ac{SNR} regime an incorrect selection always consists in an overestimation and thus we can express the probability of correct model selection as
\begin{align} \label{eq:PkPover}
\begin{array}{ccr}
\Pc\simeq1-\Pover & \quad \text{(high SNR regime)}.
\end{array}
\end{align}

\subsection{Design of the penalty}
\label{sec:design}

Theoretical and experimental results show that the probability of correct selection, $\Pc$, exhibits a sigmoidal dependence on the \ac{SNR}, raising from zero to a maximum value \cite{ZhaWon:89,Nad:10,Dju:98}. 
In particular, it has been noted that \ac{BIC} does not provide overestimations, allowing to reach a probability of correct selection close to $1$ for high \ac{SNR}. For the \ac{AIC}, instead, the maximum $\Pc$ is smaller, but it is reached at lower \acp{SNR}. This behavior of the \ac{AIC} and \ac{BIC}, reported in previous literature (see, e.g., \cite{ZhaWon:89,Kav:87,FisGroMes:02,Dju:96,Dju:98}) and confirmed by numerical results in Section~\ref{sec:numerical_results}, suggests that $\PAICk$ is too low to ensure a high $\Pc$, while $\PBICk$ is excessively high, providing good results only for high \ac{SNR}.\footnote{Note that these considerations are limited to the problems in which \ac{BIC} has been proven to provide a consistent model order selection. For example, considering the estimation of the number of signals discussed in Section~\ref{sec:WK}, this holds in presence of white noise, while it has been demonstrated that in presence of colored noise \ac{BIC} is no more a consistent model order estimator \cite{LiaReg:01,XuKav:95}. Our analysis suggests that in this case an increase in the penalty is required to compensate the dispersion of the noise eigenvalues. This problem is out of the scope of the paper and will be object of further investigations.\label{foot:BIC}}

The dependence of $\Pc$ on $\nu$ can be better understood analyzing the two error probabilities, $\Pover$ and $\Punder$, separately. Considering that $\Pkk{k}$ is always an increasing function of $k$, from the model order selection rule defined by \eqref{eq:kestITCb} and \eqref{eq:kestITC} it is easy to see that by increasing the penalty the selection of a small model order is favored, and thus a higher $\Punder$ is provided. On the other hand, when the penalty decreases a higher $\Pover$ occurs. Thus, the choice of the penalty implies a tradeoff between $\Pover$ and $\Punder$. This behaviour is confirmed by the simulation results in Section~\ref{sec:numerical_results}. 

Based on these considerations we propose to use \ac{GIC} setting the parameter $\nu$ to minimize $\Punder$ uniformly over all \acp{SNR} while $\Pover$ is constrained below a maximum value $\PoverMAX$. Note that this approach is analogous to the Neyman-Pearson criterion in binary hypothesis testing, in which $\Punder$ and $\Pover$ play the role of the probability of misdetection and the probability of false alarm, respectively. 
Considering the performance tradeoff between underestimation and overestimation, minimizing $\Punder$ corresponds to the maximization of $\Pover$, and thus the optimal value of $\nu$ is given by 
\begin{align}  \label{eq:nutildezero}
\nutilde &= \arg \max_\nu \left\{\Pover \left| \Pover \leq \PoverMAX \right.\right\}.
\end{align}
Then, since $\Pover$ is dependent on the true model order $\q$ and the \ac{SNR}, we consider a worst case design where maximization with respect to these parameters is considered. Therefore the design rule becomes
\begin{align}  \label{eq:nutildea}
\nutilde &= \arg \max_\nu \max_{q,\SNR} \left\{\Pover(\q,\SNR,\nu) \left| \Pover \leq \PoverMAX \right.\right\}\\
 &= \arg \max_\nu \max_{q} \left\{\Pover(\q,\infty,\nu)  \left| \Pover \leq \PoverMAX \right.\right\}.
 \label{eq:nutilde}
\end{align}
Equation \eqref{eq:nutilde} is due to the fact that the maximum $\Pover$ always occurs in the high \ac{SNR} regime ($\SNR\to\infty$). Given \eqref{eq:PkPover}, the approach \eqref{eq:nutilde} is equivalent to design the \ac{GIC} penalty for reaching a target probability of correct selection $\PcDES=1-\PoverMAX$ for 
high \ac{SNR}. 
Note, however, that for any \ac{SNR} it is not possible to find a $\nu<\nutilde$ that gives a lower $\Punder$, while satisfying $\max \Pover \leq \PoverMAX$.

If an analytical form for $\Pover$ is not available, we can design $\nu$ considering an upper bound on the probability of overestimation, which gives 
\begin{align}  \label{eq:nutildeLB}
 \nutilde &= \arg \max_\nu  \max_{q,\SNR}  \left\{ \PoverUB(q,\SNR,\nu) \left| \PoverUB \leq \PoverMAX \right.\right\}\nonumber\\
 &= \arg \max_\nu  \max_{q} \left\{\PoverUB(\q,\infty,\nu) \left| \PoverUB \leq \PoverMAX \right.\right\}.
\end{align}
In general, the adoption of bounds leads to a performance loss in terms of \ac{SNR}, which is smaller as the bound is tighter.

In the next sections, we discuss two examples of model order selection problems with the design of the \ac{GIC} penalty. In particular, considering the estimation of the number of signals, in Section~\ref{sec:WK} we adopt a design based on  \eqref{eq:nutilde}, while for the \ac{GLM} problem, in Section~\ref{sec:GLM}, we adopt a design based on \eqref{eq:nutildeLB}.

\section{Estimating the number of signals}
\label{sec:WK}

The problem of estimating the number of signals arises in many statistical signal processing and time series analysis applications\cite{WaxKai:85,ZhaKriBai:86,XuKav:95,FisGroMes:02,ChiWin:10,Nad:10,Kay:98,Bro:09}. We adopt the standard model in which the observation is the output of $\p$ sensors, represented, at the $i$-th time instant, by the vector
\begin{align} \label{eq:yi}
\xvi = \Hm\,\zvi+\nvi
\end{align}
where $\zvi\in\Cv{\q\times1}$ is the vector of the samples of the $\q$ signals present, $\Hm\in\Cv{\p\times\q}$ is a deterministic unknown channel matrix, and $\nvi\in\Cv{\p\times1}$ represents noise.
We assume that
$\nvi\sim\CN{\mathbf{0}}{\sig\,\Id{\p}}$, where $\sig$ is the noise power at each sensor,
and that $\zvi\sim\CN{\mathbf{0}}{\R}$.
Thus, for a given $\Hm$, the vectors $\xvi$ are zero mean Gaussian random vectors with covariance matrix
\begin{align}
\Sig=\EX{\xvi \hmt{\xvi}}=\Hm \R \hmt{\Hm} + \sig \Id{\p}.
\end{align}
Assuming that $\R$ is non singular and that the matrix $\Hm$ is of full column rank (implying $\p>\q$), which means that its columns are linearly independent vectors, the rank of $\Sig$ is $\q$ and thus the smallest $\p-\q$ eigenvalues are all equal to $\sig$.
In \cite{WaxKai:85} the estimation of the number of signals $\q$ has been posed as a model order selection problem, solved by means of \ac{ITC}. In this case we have $\p$ possible models, where the $k$-th corresponds to the situation in which exactly $k$ signals are present, with $k\in\left\{0,\dots,\p-1\right\}$. 

In this problem the parameter vector under the $k$-th hypothesis is
\begin{align}
\thetak=\left(\lambda_{\text{1}},\dots,\lambda_{k},\vv{1},\dots,\vv{k},\sig\right)
\end{align}
where $\lambda_{1}\geq\lambda_{2}\geq\dots\geq\lambda_{k}$ are the eigenvalues of $\Sig$ and $\left\{\vv{i}\right\}_{i=1,\dots,k}$ are the corresponding eigenvectors. Considering the orthonormality constraints on the eigenvectors, the number of free parameters in $\thetak$ is $\phik=k \left(2\p-k\right)+1$ \cite{WaxKai:85}.
Using the joint \ac{ML} estimates of the eigenvalues and eigenvectors of $\Sig$, obtained by \cite{And:B03}, the \ac{ITC} model order estimate \eqref{eq:kestITC} is 
\begin{align} \label{eq:waxkailath}
\qest = \arg \min_{k} \left\{-2\ln{\left(\frac{\prod_{i=k+1}^{\p} \li^{1/(\p-k)}}{\frac{1}{\p-k}\sum_{i=k+1}^{\p}\li}\right)^{\!(\p-k)\n}} + \Pkk{k} \right\}
\end{align}
where $l_1\geq l_2\geq\dots\geq l_{\p}$ are the eigenvalues of the \ac{SCM}, $\SCM=\frac{1}{\n}\Y\hmt{\Y}$, and $\Y$ is defined as \eqref{eq:Y}.

This approach is known to provide good selection performance for sufficiently large number of observations $\n$ \cite{ZhaWon:89,LiaReg:01}. For small sample sizes using the exact marginal distribution of the eigenvalues of the \ac{SCM} (without eigenvectors) gives better results \cite{ChiWin:10}. In this paper we use \eqref{eq:waxkailath} for ease of analysis.

In the following we focus on the probability of overestimation, useful for the design approach described in Section~\ref{sec:design}. Note that characterizing $\Pover$ is in general a mathematically difficult problem even in the high \ac{SNR} regime.

\subsection{Probability of overestimation}
\label{sec:Pover}

Previous works showed that in this problem for the analysis of overestimation and underestimation it is sufficient to consider the minimum of $\ITCkk{k}$ for $k=\left\{\q, \q\!\pm\!1\right\}$, which is equivalent to keep just the first term in \eqref{eq:Pover_general} and \eqref{eq:Punder_general} \cite{LiaReg:01,FisGroMes:02,Kav:87,ZhaWon:89,XuKav:95}. 
Thus, a good approximation for the probability of overestimation is given by\footnote{Note that for this problem $\PoverLB$ is a tight bound and has been often used as an approximation of $\Pover$ \cite{LiaReg:01,FisGroMes:02,Kav:87,ZhaWon:89,XuKav:95}.} 
\begin{align} \label{eq:Pover}
\Pover &\simeq \mathbb{P}\Big(\ITCkk{\q\!+\!1}<\ITCkk{\q}\!\Big) = \PoverLB.
\end{align} 
Substituting \eqref{eq:waxkailath} in \eqref{eq:Pover} 
after some manipulations we obtain the expression \cite{XuKav:95,LiaReg:01}
\begin{align} \label{eq:Pover1}
\Pover \simeq \Prob{\v \left(1-\v\right)^{\p-\q-1} < \xiq}
\end{align}
where
\begin{align} \label{eq:y}
\v = \frac{\lkk{\q+1}}{\sum_{i=\q+1}^{\p}\li}
\end{align}
and
\begin{align} \label{eq:xiq}
\xiq = \frac{\left(\p-\q-1\right)^{\p-\q-1}}{\left(\p-\q\right)^{\p-\q}} \exp\left(\frac{\Pkk{\q}-\Pkk{\q\!+\!1}}{2 \n}\right)\!.
\end{align}
The equation $\v \left(1-\v\right)^{\p-\q-1} = \xiq$ has a single real root, denoted by $\vroot$, in $[1/(\p-\q), 1]$, which is the range of $\v$. This root can be easily computed using standard root finding algorithms.\footnote{Alternaltively, an approximation of $\vroot$ is given in \cite{LiaReg:01}, while \cite{XuKav:95} provides an asymptotic expression.} 
Thus \eqref{eq:Pover1} can be expressed as
\begin{align} \label{eq:Pover3}
\Pover = 1 - F_{\v}(\vroot).
\end{align}
In the following we derive an approximated form for the computation of $F_{\v}(\cdot)$ that can be adopted for the design of $\nu$ using \eqref{eq:nutilde} and \eqref{eq:Pover3}.

\subsection{Distribution of $\v$}
\label{sec:v_distrib}

The statistic of $\v$ has been studied in \cite{ZhaWon:89,Nad:10,XuKav:95}, considering that asymptotically, for large $\n$, the smallest $\p-\q$ eigenvalues of the \ac{SCM} are distributed as the eigenvalues of a central Wishart matrix $\mathbf{W}$ with covariance matrix $\sig\, \Id{\pp}$, where $\pp=\p-\q$. 
Thus, the probability of overestimation has been evaluated considering that
\begin{align} \label{eq:vtou}
\v\stackrel{d}{\approx}\u
\end{align}
where $\u=\lone/\t$, $\lone$ and $\t$ are the largest eigenvalue and the trace of $\mathbf{W}$, respectively. In \cite{ZhaWon:89} an infinite series expression for the computation of $\Pover$ has been derived, while \cite{XuKav:95} provides an upper bound. Note that 
\eqref{eq:vtou} allows to derive an expression of $\Pover$ that is independent of the \ac{SNR}. In \cite{Nad:10} an approximation of the \ac{CDF} of $\u$ based on the Tracy-Widom distribution has been adopted. In the following we provide an approximated form of $F_{\u}(\cdot)$ that is easily invertible and is therefore useful for the design approach described in Section~\ref{sec:designWK}. Our approximation is based on the method of moments, which consists in choosing a simple distribution model and setting its parameters to match the first exact moments \cite{MarGioChi:12,GioChi:13,Chi:14}. 

As shown in \cite{Mui:82}, the moments of $\u$ can be computed considering that, conditioned on $\lone$, $\u$ and $\t$ are independently distributed, which leads to
\begin{align} \label{eq:moments_u}
\mk{i}=\mlonek{i}/\mtk{i}
\end{align}
where $\mk{i}$, $\mlonek{i}$ and $\mtk{i}$ are the $i$-th moments of $\u$, $\lone$ and $\t$, respectively.

In \cite{ZanChiWin:09} and \cite{ZanChi:12} it has been shown that the \ac{p.d.f.} of $\lone$ is a gamma mixture distribution, which can be expressed as a linear combination of gamma-shaped functions as
\begin{align}\label{eq:flone}
f_{\lone}(x)=\sum_{s=1}^{\pp}\sum_{j} \,\epsilon_{s,j} \, x^{j} \, e^{-s x}.
\end{align}
The evaluation of the parameters $\epsilon_{s,j}$ can be found in \cite{ZanChiWin:09} and \cite{ZanChi:12}. Based on \eqref{eq:flone} the $i$-th moment of $\lone$ can be derived in closed-form as
\begin{align} \label{eq:moments_lone}
\mlonek{i}&=\int_0^{\infty} x^{i}\,f_{\lone}(x)dx\nonumber\\
& = \sum_{s=1}^{\pp}\sum_{j} \frac{\epsilon_{s,j}}{{s}^{j+i+1}}\,\Gammaf{j+i+1}
\end{align}
where $\Gammaf{a}\triangleq\int_{0}^{\infty}y^{a-1} e^{-y} dy$ is the gamma function. Alternative methods for computing the moments of $\lone$ based on integral expressions or approximations are discussed in Appendix A.

Considering that $\t\sim\mathcal{G}\!\left(\n\,\pp,1\right)$ \cite{Mui:82}, the moments of the trace are given by
\begin{align} \label{eq:moments_t}
\mtk{i}=\frac{\Gammaf{\pp\n+i}}{\Gammaf{\pp\n}}.
\end{align}

\begin{figure}[t]
%
\psfrag{cdf}{\scriptsize $F_{\u}(x)$}
\psfrag{x}{\scriptsize $x$}
\psfrag{n10}{\hspace{-0.38cm}\scriptsize $\n=10$}
\psfrag{n20}{\hspace{-0.38cm}\scriptsize $\n=20$}
\psfrag{p8}{\hspace{-0.38cm}\scriptsize $\p=8$}
\psfrag{p3}{\scriptsize $\p=3$}
\psfrag{n8}{\scriptsize $\n=8$}
\psfrag{Est}{\scriptsize simulated}
\psfrag{App}{\scriptsize approximated}
\psfrag{Exact}{\scriptsize exact}
%
%
\centering
\includegraphics[width=0.9\columnwidth,draft=false,clip]{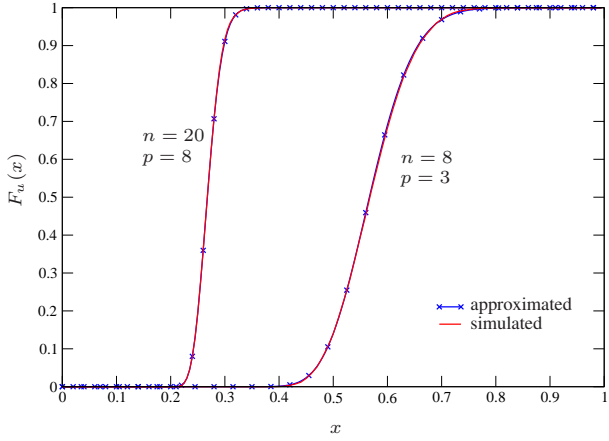}
\caption[]{Comparison among the simulated \ac{CDF} of $\u$ and the approximated form in \eqref{eq:cdf_u_appr}.}
\label{fig:cdf_u}
\end{figure}

Once the moments $\mk{i}$ are computed using \eqref{eq:moments_u}, \eqref{eq:moments_lone} and \eqref{eq:moments_t}, we approximate $\u$ to a shifted gamma distributed \ac{r.v.} as 
\begin{align}
\u + \alpha \stackrel{d}{\approx}\mathcal{G}\!\left(\kappa,\theta\right)
\end{align}
where $\kappa$, $\theta$, and the shift $\alpha$ are expressed as \cite{Chi:14}
\begin{align}
\kappa&=\frac{4\,\left(\mk{2}-\mk{1}^2\right)^3}{\left(\mk{3} - 3 \mk{1} \mk{2} + 2 \mk{1}^3\right)^2}\\
\theta&=\frac{\mk{3} - 3 \mk{1} \mk{2} + 2 \mk{1}^3}{2 \left(\mk{2}-\mk{1}^2\right)}\\
\alpha&=\kappa\, \theta - \mk{1}.
\end{align}
Thus the approximated \ac{CDF} of $\u$ is given by
\begin{align} \label{eq:cdf_u_appr}
F_{\u}(x)\simeq\begin{cases}{}
\IncLowerGammaf{\kappa}{\frac{x+\alpha}{\theta}}, & x > -\alpha\\
0, & x\leq-\alpha
\end{cases}
\end{align}
where $\IncLowerGammaf{a}{z}\triangleq\frac{1}{\Gammaf{a}}\int_{0}^{z}y^{a-1} e^{-y} dy$ is the normalized incomplete gamma function. Note that \eqref{eq:cdf_u_appr} can be inverted using the inverse incomplete gamma function, which is already implemented in standard mathematical software. In Fig.~\ref{fig:cdf_u} the comparison between the simulated and approximated \ac{CDF} of $\u$ are reported. As can be seen, the shifted gamma approximation in \eqref{eq:cdf_u_appr} matches very well the simulated distribution of $\u$. 

\subsection{Design of the penalty}
\label{sec:designWK}

The \ac{GIC} penalty can be now designed using \eqref{eq:nutilde} and the expression of $\Pover$ in \eqref{eq:Pover3}. Using the relation
\begin{align} \label{eq:design2}
\max_q \Pover = \left.\Pover\right|_{\q=\qstar} \simeq 1 - F_{\left.\u\right|_{\q=\qstar}}(\v)
\end{align}
where $\qstar=\arg\max_{\q} \Pover$,
we obtain the value of $\v$ that corresponds to $\max_{\q} \Pover = \PoverMAX$ as
\begin{align}
\vtildeqstar = F_{\left.\u\right|_{\q=\qstar}}^{-1}\!\!\left(1-\PoverMAX\right).
\end{align}
Then the optimal $\nu$ to be used in \eqref{eq:PkGIC} is given by 
\begin{align} \label{eq:nutildeWK}
\nutilde =& - \frac{2 \n}{2\left(\p-\qstar\right)-1} \ln \left(\frac{(\p-\qstar)^{\p-\qstar}}{(\p-\qstar-1)^{\p-\qstar-1}}\right)\nonumber\\
& \times \ln\left(\vtildeqstar \left(1-\vtildeqstar\right)^{\p-\qstar-1}\right).
\end{align}
Numerical results assessing the effectiveness of this design strategy are presented in Section~\ref{sec:numerical_results_WK}.

\section{General linear model}
\label{sec:GLM}

The \ac{GLM} can be applied to a large set of problems in different fields of science and engineering (see \cite{Gra:76} and \cite{Dju:98} for some examples).
Under the \ac{GLM}, the observation consists in a $\n$ length random vector defined as\footnote{In this case the samples are scalars ($\p=1$), and thus the observation matrix $\Y$ in \eqref{eq:Y} reduces to the row vector $\yvv$, and the vectors $\xvi$ reduce to the scalars $x_i$.} 
\cite{Gra:76}
\begin{align} \label{eq:yGLM}
\yvv = \transp{\betavv}\,\Hk{\q}  + \nvv
\end{align}
where $\Hk{q}$ is a $\psiq\times\n$ matrix of known fixed values with linearly independent columns, $\betavv\in\Cv{\psi\left(q\right)\times1}$ 
is a vector of unknown deterministic parameters, and $\transp{\nvv}\sim\CN{\mathbf{0}}{\sig\,\Id{\n}}$.\footnote{We assume that the observed samples are complex \acp{r.v.}. The analysis of the real case, adopted e.g. in \cite{Dju:98}, can be derived as a special case.} 
%
%
In this case, the selection problem consists in estimating the length of $\betavv$, $\psiq$, which is a function of the model order $\q$. 
Here, \ac{ITC} can be applied using  \eqref{eq:kestITCb}, \eqref{eq:kestITC}, and the fact that under the $k$-th hypothesis we have \cite{Dju:98}
\begin{align}
\label{eq:likelihood_glm} 
-2 \sum_{i=1}^{\n} \lnfcond{x_i}{\thetakest} = \n \ln \sigestk
\end{align}
where
\begin{align}
\label{eq:likelihood_glm2} & \sigestk = \frac{1}{\n} \yvv \Mproj{k} \hmt{\yvv}
\end{align}
and
\begin{align}
\label{eq:projmatrix} & \Mproj{k} = \Id{\n} - \HkH{k} \left(\Hk{k} \HkH{k}\right)^{\!-1}\! \Hk{k}
\end{align}
is a projection matrix.

\subsection{Bounds on the probability of overestimation}
\label{sec:GLMbounds}

Differently from the case in the previous section, being the derivation of an analytic expression of \eqref{eq:Pover_general} non trivial, we adopt the bound based approach described in Section~\ref{sec:design}. 
The probability $\Prob{\ITCkk{q+i}<\ITCkk{q}}$ can be expressed, using \eqref{eq:kestITCb}, \eqref{eq:likelihood_glm} and \eqref{eq:likelihood_glm2}, as
\begin{align} \label{eq:probRi}
\Prob{\ITCkk{q+i}<\ITCkk{q}} = \Prob{R_i < \exp\left(-\frac{2 i \nu}{\n}\right)}
\end{align}
where
\begin{align} \label{eq:Ri}
R_i =\frac{\yvv \Mproj{\q+i} \hmt{\yvv}}{\yvv \Mproj{\q} \hmt{\yvv}}.
\end{align}
In the Appendix B we prove that $R_i$ is a beta distributed \ac{r.v.} with parameters $\n-2\left(\q-i\right)$ and $2i$, and thus the terms in \eqref{PoverUB} and \eqref{PoverLB} can be expressed in closed-form as
\begin{align} \label{eq:probR}
\Prob{\ITCkk{q+i}<\ITCkk{q}}= \BetaIf{\exp\left(-2 i \nu/\n\right)}{\n-2\left(\q-i\right)}{2i}
\end{align}
where $\BetaIf{x}{a}{b}=\frac{1}{\Betaf{a}{b}}\int_{0}^{x}z^{a-1}(1-z)^{b-1}dz$, with $0\leq x\leq1$, is the incomplete beta function and $\Betaf{a}{b}=\Gammaf{a}\,\Gammaf{b}/\Gammaf{a+b}$ is the beta function. Thus, by using \eqref{eq:probR}, we can easily compute $\PoverUB$ in \eqref{PoverUB}. Note, in particular, that the probability in \eqref{eq:probRi}, and thus also the bounds \eqref{PoverUB} and \eqref{PoverLB}, does not depend on the \ac{SNR}.

\subsection{Design for the GLM}
\label{sec:GLMdesign}

Based on the bound derived, the design of the penalty can be performed according to \eqref{eq:nutildeLB}. 
In this problem the probability in \eqref{eq:probR} is a decreasing function of $\q$, and thus the maximum in \eqref{eq:nutildeLB} is reached for $\q=0$, giving
\begin{align}  \label{eq:nutildeGLM}
\nutilde = \arg \max_\nu \left\{\PoverUB(0,\infty,\nu) \left| \PoverUB \leq \PoverMAX \right. \right\}
\end{align}
which can be numerically computed inverting \eqref{PoverUB}. Numerical results based on \eqref{eq:nutildeGLM} are presented in Section~\ref{sec:numerical_results_GLM}.

Note that when $\q=0$ we cannot have underestimation, and thus \eqref{eq:PkPover} holds in general, not only for high \ac{SNR}. Moreover, note that when $\q=0$, $\Pover$ corresponds to the probability that model selection fails when only noise is present, i.e., the probability of false alarm in signal detection. Therefore, in this case our design strategy corresponds to the Neyman-Pearson design criterion, in which the target probability of false alarm is $\PoverMAX$.

\section{Numerical results}
\label{sec:numerical_results}

In this section we present some numerical results to prove the effectiveness of the design approach proposed.

\begin{figure}[t]
%
%
\psfrag{pok}{\scriptsize $\Pc$}
\psfrag{SNR}{\scriptsize \hspace{-0.2cm}$\SNR$ [dB]}
\psfrag{GIC20}{\tiny \ac{GIC}, $\nu=2$ (AIC)}
\psfrag{GIC22}{\tiny \ac{GIC}, $\nu=2.2$}
\psfrag{GIC25}{\tiny \ac{GIC}, $\nu=2.5$}
\psfrag{GIC30}{\tiny \ac{GIC}, $\nu=3$}
\psfrag{BIC}{\tiny \ac{GIC}, $\nu=\ln\n$ (BIC)}
\centering
\includegraphics[width=0.9\columnwidth,draft=false,clip]{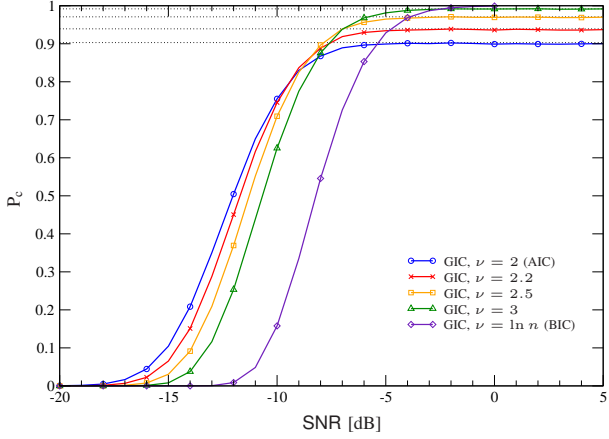}
\caption[]{Probability of correct model order selection as function of the \ac{SNR} for the problem of estimating the number of signals in \ac{AWGN} when $\q=4$, $\p=8$, $\n=1000$. The maximum $\Pc$ (dotted lines) is approximated using \eqref{eq:Pover3} and \eqref{eq:cdf_u_appr}.}
\label{fig:pokGIC}
\end{figure}
\begin{figure}[t]
%
\psfrag{GIC1}{\tiny \ac{GIC}, $\nu=2$ (AIC)}
\psfrag{GIC2}{\tiny \ac{GIC}, $\nu=2.2$}
\psfrag{GIC3}{\tiny \ac{GIC}, $\nu=2.5$}
\psfrag{GIC4}{\tiny \ac{GIC}, $\nu=3$}
\psfrag{BIC}{\tiny \ac{GIC}, $\nu=\ln\n$ (BIC)}
\psfrag{pfa}{\scriptsize $\Pover$}
\psfrag{pmd}{\scriptsize$\Punder$}
\psfrag{pfapmd}{\scriptsize $\Pover$, $\Punder$}
\psfrag{SNR}{\scriptsize \hspace{-0.2cm}$\SNR$ [dB]}
%
%
\centering

\subfigure[Probability of overestimation]{
\includegraphics[width=0.9\columnwidth,draft=false,clip]{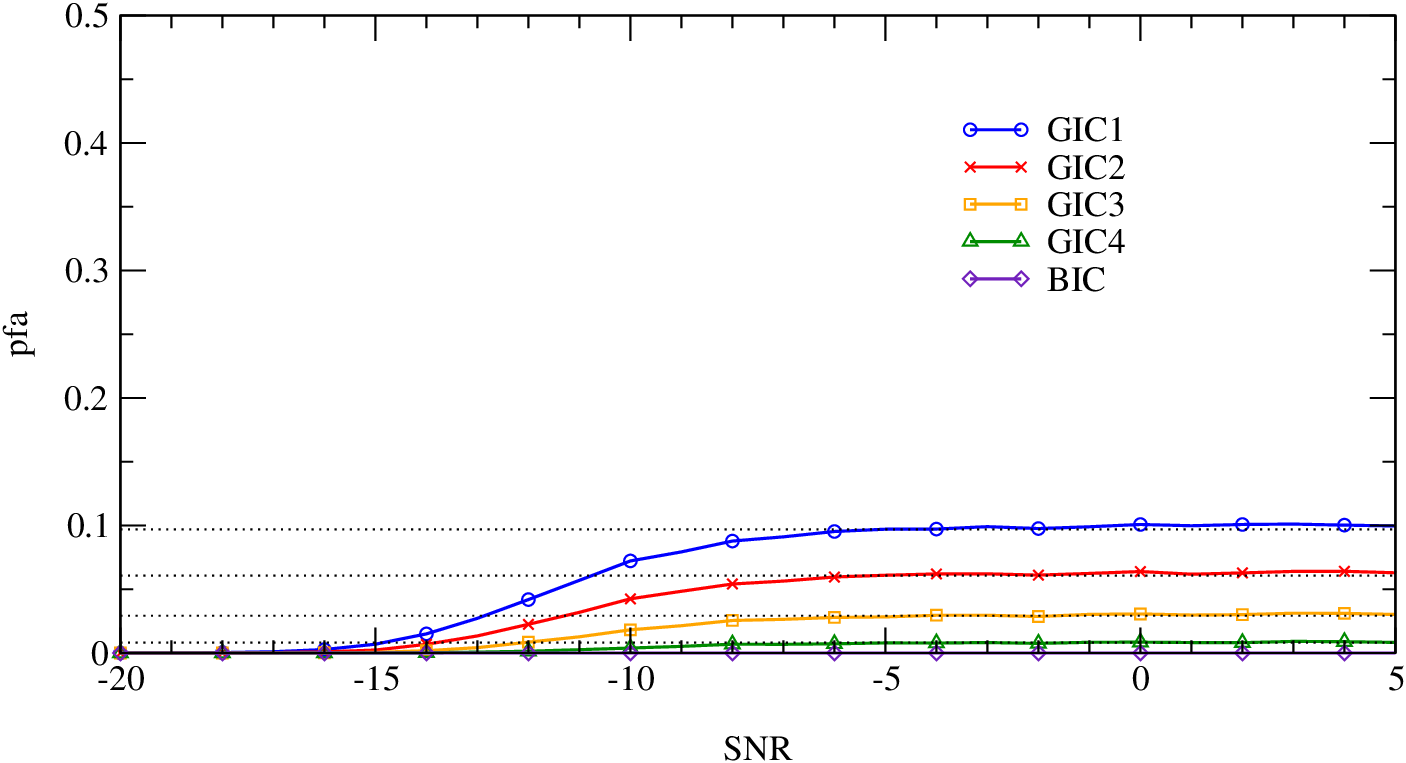}
\label{fig:pfaGIC}
}

\subfigure[Probability of underestimation]{
\includegraphics[width=0.9\columnwidth,draft=false,clip]{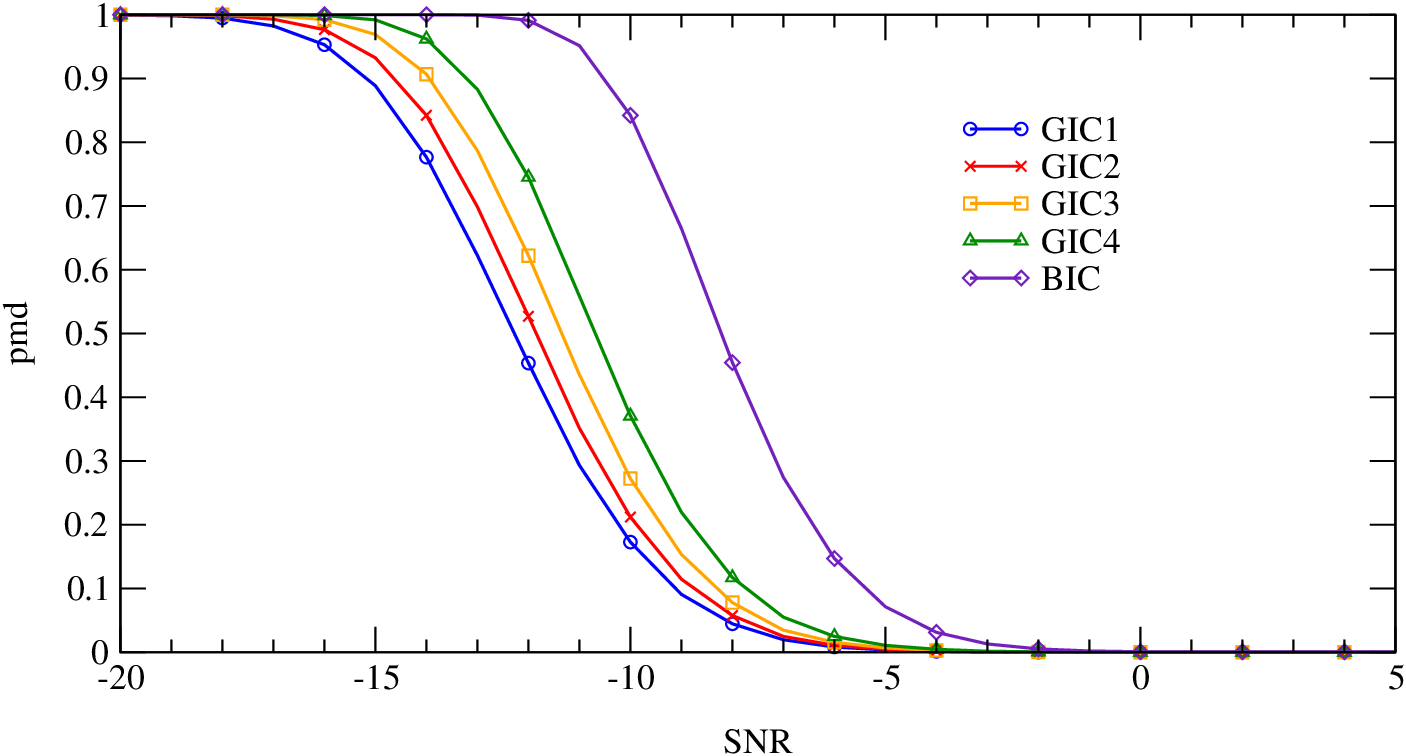}
\label{fig:pmdGIC}
}
\caption[]{Probabilities of model order overestimation and underestimation as function of the \ac{SNR} considering $\q=4$ signals in \ac{AWGN} when $\p=8$, $\n=1000$. The maximum probability of overestimation (dotted lines) is approximated using \eqref{eq:Pover3} and \eqref{eq:cdf_u_appr}.}
\label{fig:pfapmdGIC}
\end{figure}

\subsection{Estimating the number of transmitting sources}
\label{sec:numerical_results_WK}

Considering the problem described in Section~\ref{sec:WK}, we focus, as an example, on the estimation of the number of transmitting sources by a multiple antenna system that arises in array signal processing and cognitive radio contexts \cite{ChiWin:10,WaxKai:85}. Thus in this case $\xvi$ is the vector of the output samples of the sensor antennas at the $i$-th time instant after downconversion and sampling, $\svi$ is the vector of the samples of the $\q$ signals present, $\Hm$ describes the gain of the radio channel between the $\q$ signal sources and the $\p$ antennas, and $\nvi$  represents the thermal noise. For this problem we have $\SNR=\tr{\Hm \R \hmt{\Hm}}/(\p\,\sig)$. 

In Fig.~\ref{fig:pokGIC} we show $\Pc$ as function of the \ac{SNR} when $\q=4$, $\p=8$ and $\n=1000$. We can see that the curves confirm the behaviour described in Section~\ref{sec:perfdesign}. Considering \ac{AIC}, we can see that it reaches a maximum $\Pc$ of about $0.9$, while \ac{BIC} provides probability of correct selection almost one at the expense of a loss for $\SNR<-3\,$dB. By changing the \ac{GIC} parameter $\nu$ we can trade-off between the high and low \ac{SNR} performance. Note that the maximum $\Pc$ is correctly predicted using \eqref{eq:nutilde}, \eqref{eq:Pover3} and \eqref{eq:cdf_u_appr} (dotted lines). The corresponding overestimation and underestimation probabilities are shown in Fig.~\ref{fig:pfapmdGIC}. We can see that an increase of $\nu$ gives a lower $\Pover$ but a higher $\Punder$. Note, in particular, that by increasing the \ac{SNR} $\Punder$ goes to zero, which supports the approximation in \eqref{eq:PkPover}. Also note that \eqref{eq:PkPover} is a very {favorable} property in {CR} scenarios, implying that \ac{ITC} never misdetect the presence of \acp{PU} if the \ac{SNR} is sufficiently high.

\begin{figure}[t]
\psfrag{pok}{\scriptsize $\Pc$}
\psfrag{pfa}{\scriptsize $\Pover$}
\psfrag{k}{\scriptsize $\q$}
\psfrag{GIC1}{\tiny $\nu=1.7$}
\psfrag{GIC2}{\tiny $\nu=1.8$}
\psfrag{GIC3}{\tiny $\nu=1.9$}
\psfrag{GIC4}{\tiny $\nu=2$}
\psfrag{GIC5}{\tiny $\nu=2.2$}
\psfrag{GIC6}{\tiny $\nu=2.5$}
\psfrag{GIC7}{\tiny $\nu=3$}
\psfrag{GIC8}{\tiny $\nu=3.5$}
\centering
\includegraphics[width=0.9\columnwidth,draft=false,clip]{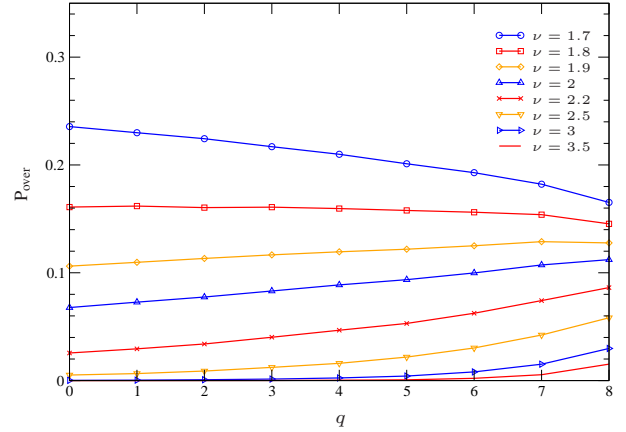}
\caption[]{Probability of overestimation as function of $\q$ for the problem of estimating the number of signals in \ac{AWGN}, for different values of the \ac{GIC} parameter $\nu$, and $\p=10$, $\n=1000$, $\SNR=5\,$dB.}
\label{fig:pokGICk}
\end{figure}

\begin{figure}[t]
\psfrag{pok}{\scriptsize $\Pc$}
\psfrag{SNR}{\scriptsize \hspace{-0.2cm}$\SNR$ [dB]}
\psfrag{q0}{\tiny $\q\!=\!0$}
\psfrag{q1}{\tiny $\q\!=\!1$}
\psfrag{q2}{\tiny $\q\!=\!2$}
\psfrag{q3}{\tiny $\q\!=\!3$}
\psfrag{q4}{\tiny $\q\!=\!4$}
\psfrag{PokLB}{\tiny $1\!-\!\left.\Pover\right|_{\q=4}$}
\centering
\includegraphics[width=0.9\columnwidth,draft=false,clip]{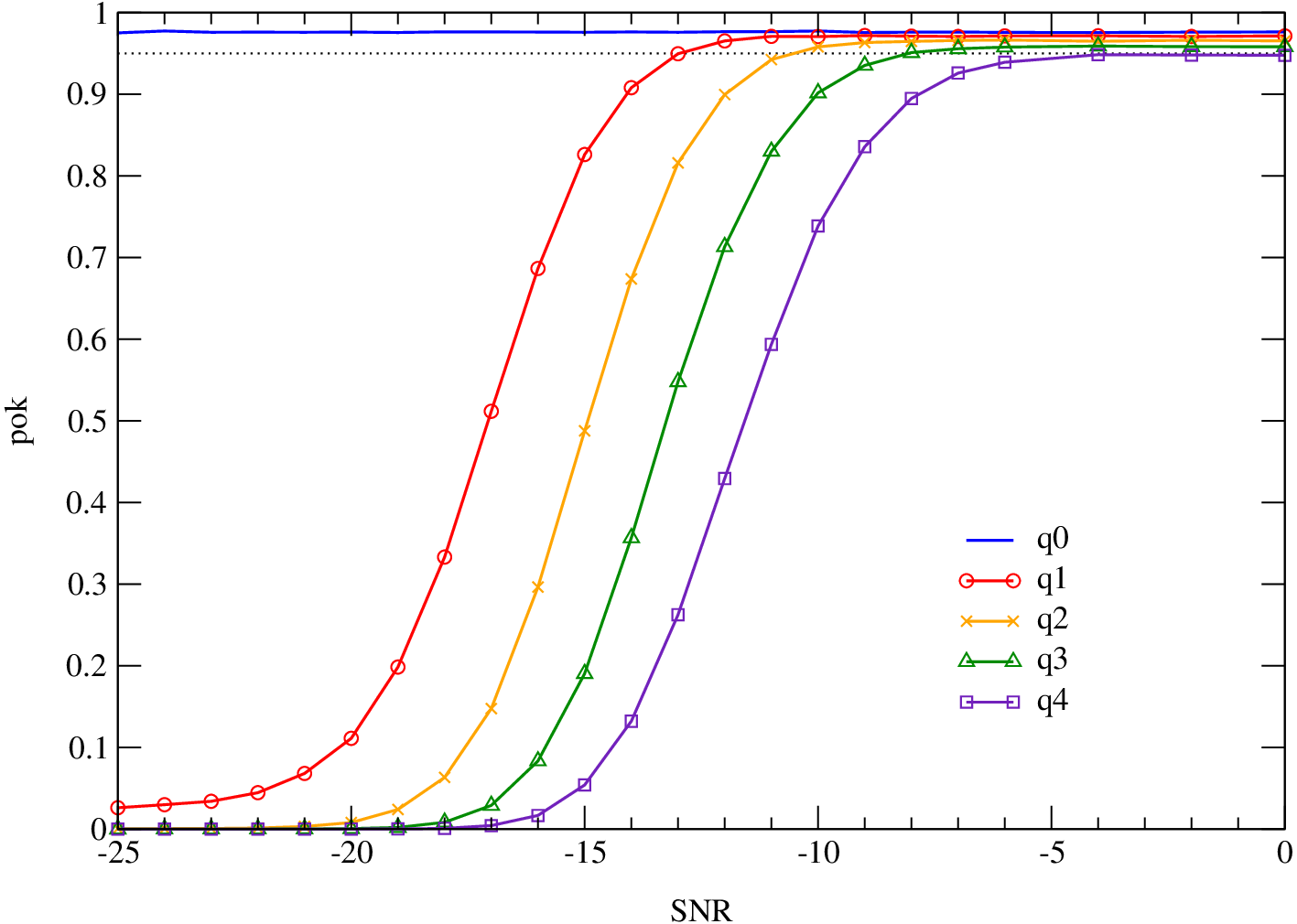}
\caption[]{Probability of correct model order selection as function of the \ac{SNR} for the problem of estimating the number of signals in \ac{AWGN} when $\nu$ is set according to \eqref{eq:nutildeWK} with $\PoverMAX=0.05$, $\qmax=4$, $\p=8$ and $\n=1000$.}
\label{fig:pokGICdesign}
\end{figure}

In Fig.~\ref{fig:pokGICk} we show $\Pc$ as a function of the number of signal sources. We can see that in different situations the maximum occurs for different $\q$ and thus, in general, the maximization in \eqref{eq:nutilde} requires the evaluation of $\Pc$ for all the number of sources 
considered. Note, however, that for high $\Pc$, e.g. greater than $0.9$, which is the most interesting case in practice, the curves decrease with $\q$, and thus the design can be based on $\qstar=\qmax$.

In Fig.~\ref{fig:pokGICdesign} we show $\Pc$ as function of the \ac{SNR} considering $\qmax=4$, $\p=10$ and $\n=1000$. Using \eqref{eq:nutildeWK} with $\PoverMAX=0.05$ we obtain $\nutilde=2.281$. Note that when $\q=\qmax$, for high \ac{SNR}, $\Pc$ coincides with $1-\PoverMAX$, while when $\q<\qmax$, we reach, as expected, a higher probability of correct selection. From the comparison with Fig.~\ref{fig:pokGIC} ($\q=4$ case) we can see that when $\SNR=0\,$dB \ac{BIC} provides probability of correct selection almost one, while \ac{AIC} gives $\Pc\approx 0.9$. Note that the advantage of \ac{BIC} is lost at lower \acp{SNR}. For example, considering $\SNR=-10\,$dB, \ac{BIC} provides $\Pc\approx 0.16$, while \ac{GIC} with the design of the penalty gives $\Pc\approx 0.76$.

\subsection{Estimating the number of sinusoids in \ac{AWGN}}
\label{sec:numerical_results_GLM}

In this section we focus, as an example of \ac{GLM}, on the problem of estimating the number of sinusoids in \ac{AWGN}, described in \cite{Dju:96,Dju:98,NadKon:11,MaNg:06}. In this case, the $i$-th element of $\yvv$ in \eqref{eq:yGLM} is given by
\begin{align}
x_i = \sum_{l=1}^{\q} \aii{l}\, e^{\,\jmath\left(2\pi\fii{l}i+\phaseii{l}\right)}+n_i
\end{align}
that can be rewritten as
\begin{align}
x_i = \sum_{l=1}^{\q} \aii{l}\,e^{\jmath \phaseii{l}} \cos\!\left(2\pi\fii{l}i\right) + \jmath\, \aii{l}\,e^{\jmath \phaseii{l}} \sin\!\left(2\pi\fii{l}i\right) + n_i.
\end{align}
where $n_i$ is the $i$-th element of $\nvv$, and $\jmath=\sqrt{-1}$.

We assume, as in \cite[Section~IV-A]{Dju:98}, that the sinusoids considered are taken from a known frequency set $\left\{\fii{k}\right\}_{k\in\mathcal{K}}$ and that, considering the $k$-th hypothesis, the matrix $\Hk{k}$ is given by \cite{Dju:98}
\begin{align} \label{eq:Xk}
\Hk{k} = \left(\hvki{1} | \hvki{2} | \dots | \hvki{\n}\right)
\end{align}
where
\begin{align}
\hvki{i} = & \left(\cos(2\pi\fii{1} i), \sin(2\pi\fii{1} i), \cos(2\pi\fii{2} i), \sin(2\pi\fii{2} i), \right. \nonumber\\
&\quad\transp{\left.\dots,  \cos(2\pi\fii{k} i), \sin(2\pi\fii{k} i) \right)}.
\end{align}
The vector $\betavv$, that contains the information on the sinusoids amplitudes and phases, has a length $\psi\!\left(k\right)=2 k$ and is given by
\begin{align}
\betavv=\transp{\left(\aii{1}e^{\jmath \phaseii{1}}\!,\;  \jmath\aii{1}e^{\jmath \phaseii{1}}\!,\; \aii{2}e^{\jmath\phaseii{2}}\!,\;  \jmath\aii{2}e^{\jmath \phaseii{2}}\!,  \; \dots\; \aii{k}e^{\jmath \phaseii{k}}\!, \; \jmath\aii{k}e^{\jmath \phaseii{k}} \right)}.
\end{align}
In this problem, the number of free parameters in the $k$-th hypothesis is $\phik=2\,k+1$, accounting for the $k$ unknown amplitudes, the $k$ unknown phases, and the noise power, and the \ac{SNR} is given by $\SNR=\sum_{l=1}^{k} |\aii{l}|^2/\sig$.

\begin{figure}[t]
\psfrag{pok}{\scriptsize $\Pc$}
\psfrag{SNR}{\scriptsize \hspace{-0.2cm}$\SNR$ [dB]}
\psfrag{AIC}{\tiny \ac{GIC}, $\nu=2$ (AIC)}
\psfrag{GIC2}{\tiny \ac{GIC}, $\nu=2.5$}
\psfrag{GIC3}{\tiny \ac{GIC}, $\nu=3$}
\psfrag{GIC4}{\tiny \ac{GIC}, $\nu=4$}
\psfrag{BIC}{\tiny \ac{GIC}, $\nu=\ln\n$ (BIC)}
\centering
\includegraphics[width=0.9\columnwidth,draft=false,clip]{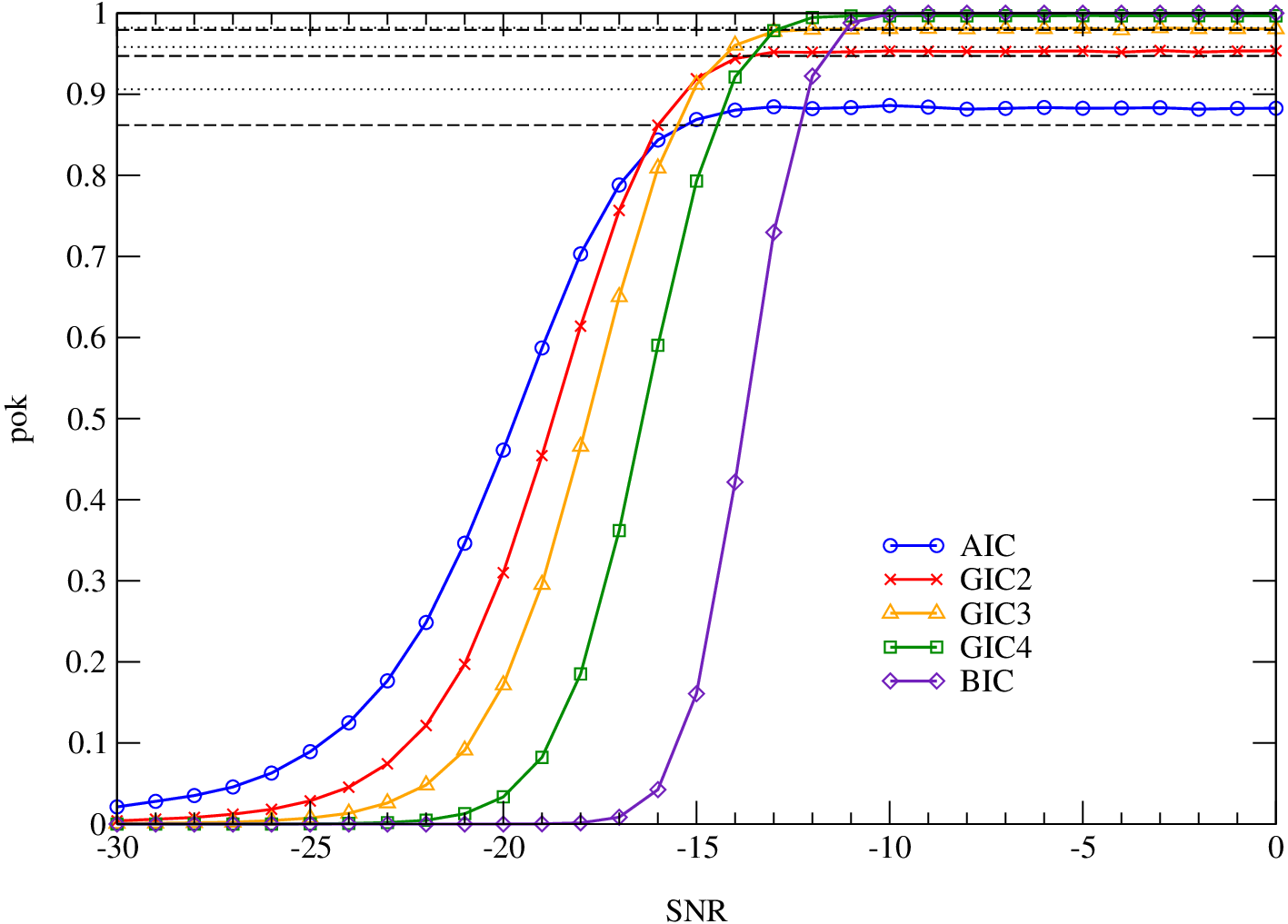}
\caption[]{Probability of correct model order selection as function of the \ac{SNR} for the problem of estimating the number of sinusoids in \ac{AWGN} when $\q=3$, $\qmax=6$, $\n=1000$. The lower and upper bounds of the maximum $\Pc$ correspond to the dashed and dotted lines, respectively.
}
\label{fig:pokGLM}
\end{figure}

\begin{figure}[t]
\psfrag{AIC}{\tiny \ac{GIC}, $\nu=2$ (AIC)}
\psfrag{GIC2}{\tiny \ac{GIC}, $\nu=2.5$}
\psfrag{GIC3}{\tiny \ac{GIC}, $\nu=3$}
\psfrag{GIC4}{\tiny \ac{GIC}, $\nu=4$}
\psfrag{BIC}{\tiny BIC}
\psfrag{pfa}{\scriptsize $\Pover$}
\psfrag{pmd}{\scriptsize$\Punder$}
\psfrag{pfapmd}{\scriptsize $\Pover$, $\Punder$}
\psfrag{SNR}{\scriptsize \hspace{-0.2cm}$\SNR$ [dB]}
\centering

\subfigure[Probability of overestimation]{
\includegraphics[width=0.9\columnwidth,draft=false,clip]{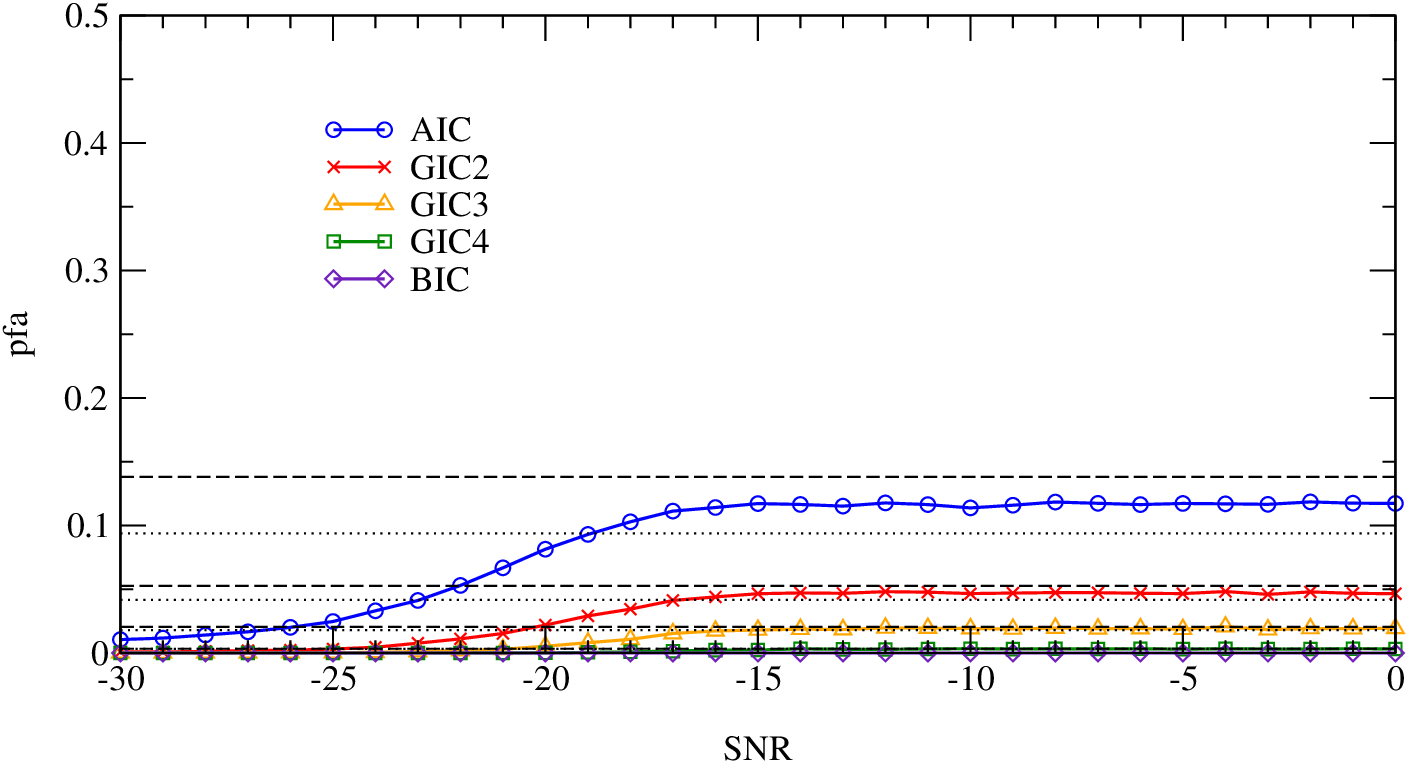}
\label{fig:pfaGIC}
}

\subfigure[Probability of underestimation]{
\includegraphics[width=0.9\columnwidth,draft=false,clip]{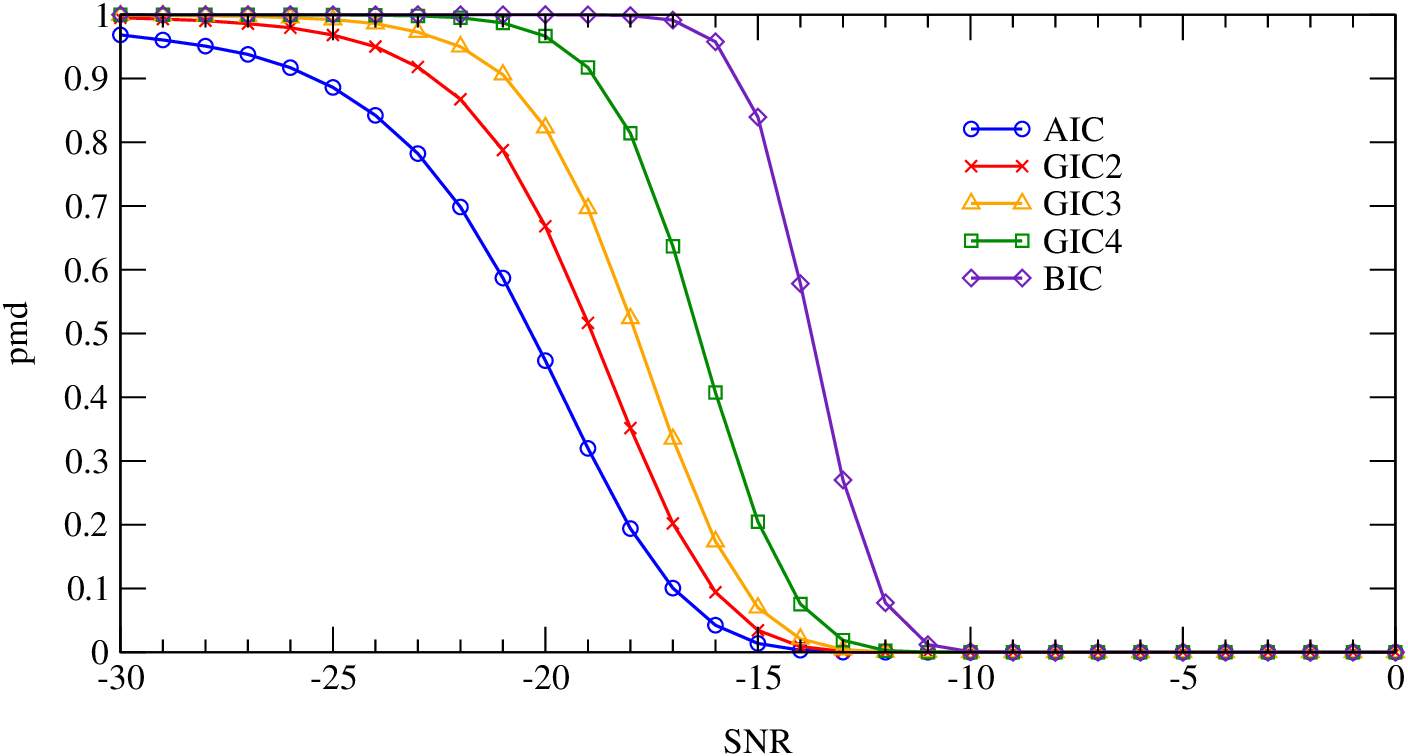}
\label{fig:pmdGIC}
}
\caption[]{Probability of incorrect model order selection as function of the \ac{SNR} for the problem of estimating the number of sinusoids in \ac{AWGN} when $\q=3$, $\qmax=6$, $\n=1000$. The lower and upper bounds of the maximum $\Pover$ correspond to the dotted and dashed lines, respectively.
}
\label{fig:pfapmdGLM}
\end{figure}

Differently from the case in the previous section, in the following we adopt the performance bounds described in Section~\ref{sec:perf}. Numerical simulations show that a good approximation for $\Pc$ is provided when $\imax=2$ in \eqref{PoverUB} and \eqref{PoverLB}. Therefore we have
\begin{align} \label{eq:PoverUB_GLM}
\PoverUB=&\BetaIf{\exp\left(-\frac{2 \nu}{\n}\right)}{\n-2\left(\q+1\right)}{2}\nonumber\\
&+\BetaIf{\exp\left(-\frac{4 \nu}{\n}\right)}{\n-2\left(\q+2\right)}{4}\\
\PoverLB=&\BetaIf{\exp\left(-\frac{2 \nu}{\n}\right)}{\n-2\left(\q+1\right)}{2}.
\label{eq:PoverLB}
\end{align}

We adopt, in particular, the example proposed in \cite{Dju:96}, in which the $k$-th frequency in the considered set is $f_k=0.2+(k-1)/\n$, with $k=1,\dots,\qmax$.

In Fig.~\ref{fig:pokGLM} we show $\Pc$ as function of the \ac{SNR} when $\q=3$, $\qmax=6$ and $\n=1000$. The three sinusoids have equal amplitude and phases $0$, $\pi/4$ and $\pi/3\,$rad, respectively. The corresponding error probabilities are plotted in Fig.~\ref{fig:pfapmdGLM}. Also for this problem we can see that the curves confirm the behaviour described in Section~\ref{sec:perfdesign}. We also plot the upper and lower bounds for $\Pover$ and the corresponding bounds for $\Pc$. Note that increasing $\Pc$ the bounds become tighter, and thus they can be considered $\Pc$ approximations.

\begin{figure}[t]
\psfrag{pok}{\scriptsize $\Pc$}
\psfrag{SNR}{\scriptsize \hspace{-0.2cm}$\SNR$ [dB]}
\psfrag{q0}{\tiny $\q=0$}
\psfrag{q1}{\tiny $\q=1$}
\psfrag{q2}{\tiny $\q=2$}
\psfrag{q3}{\tiny $\q=3$}
\psfrag{q4}{\tiny $\q=4$}
\psfrag{PokLB}{\tiny $1-\PoverUB$}
\centering
\includegraphics[width=0.9\columnwidth,draft=false,clip]{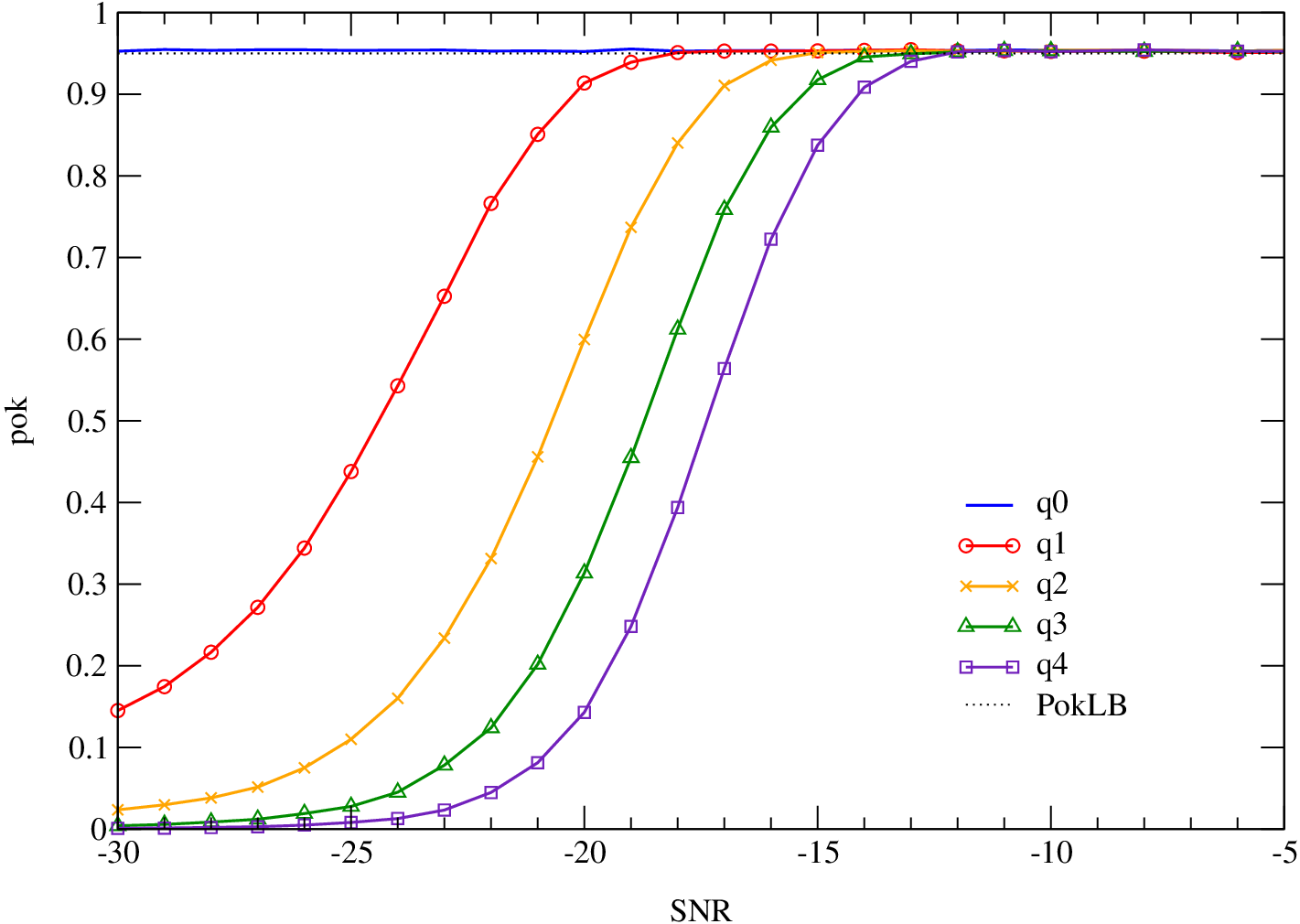}
\caption[]{Probability of correct model order selection as function of the \ac{SNR} for the problem of estimating the number of sinusoids in \ac{AWGN}, when $\nu$ is set according to \eqref{eq:nutildeLB} and \eqref{eq:PoverUB_GLM}, with $\PoverMAX=0.05$ and $\n=1000$. 
}
\label{fig:pokGLMdesign}
\end{figure}

An example of penalty design for the problem of estimating the number of sinusoids is reported in Fig.~\ref{fig:pokGLMdesign}. Choosing $\PoverMAX=0.05$ and $\n=1000$, from \eqref{eq:nutildeGLM} and \eqref{eq:PoverUB_GLM} we obtain $\nu=2.499$. We can see that the maximum $\Pc$ is always above the bound $1-\PoverUB$, which, in this case, is very tight to the estimated curves. From the comparison with Fig.~\ref{fig:pokGLM} ($\q=3$ case) we can see that for $\SNR=0\,$dB \ac{BIC} provides probability of correct selection almost one, while \ac{AIC} gives $\Pc\approx 0.89$. Note that the advantage of \ac{BIC} is lost a lower \acp{SNR}. For example, considering $\SNR=-15\,$dB, \ac{BIC} provides $\Pc\approx 0.16$, while \ac{GIC} with the design of the threshold gives $\Pc\approx 0.92$.

\section{Conclusion}
\label{sec:conclusions}

In this paper, we studied model order selection based on \ac{ITC} under a design perspective. We focused on the \ac{GIC}, which embraces most common criteria, and we proposed a strategy for designing its penalty for finite sample sizes. This method allows to keep the probability of overestimation below a specified level. We applied this design strategy to two model selection problems. Firstly, we studied the problem of estimating the number of sources, which received considerable attention in the past decades. We provided, in particular, a new approximated form for the computation of the maximum probability of correct selection based on the ratio of the largest eigenvalue to the trace of a central white Wishart matrix. We also applied model selection to the \ac{GLM}, proposing a design strategy based on the bounds of the probability of overestimation, which can be applied to any selection problem with nested hypotheses. As a particular case, we focused on the problem of estimating the number of sinusoids in \ac{AWGN}. 
In both case studies we showed that the high \ac{SNR} performance analysis can be addressed independently on the signal adopted.
The proposed design strategy aims to choose proper \ac{ITC} penalties to control the model order selection performance in finite sample size problems.



\section*{Appendix A}
\label{sec:appendixA}

In Section~\ref{sec:v_distrib} we provide the exact expression of the moments of $\lone$ based on the gamma mixture distribution \eqref{eq:flone}. In the following we propose two alternative approaches for simplifying their computation.

For large $\n$ and $\pp$, $\lone$ can be approximated using simpler distributions. For instance, a well known approximation of $\lone$ is related to the Tracy-Widom distribution \cite{Joh:01}. Recently, it has been shown that \cite[eq. (48)]{Chi:14}
\begin{align}
\frac{\lone - \mu_{np}}{\sigma_{np}} + \widetilde{\alpha} \stackrel{d}{\approx} \mathcal{G}\!\left(\widetilde{\kappa},\widetilde{\theta}\right)
\end{align}
where $\mu_{np}=\left(\sqrt{\n} +\sqrt{\pp}\right)^2$, $\sigma_{np}=\sqrt{\mu_{np}}$ $\times \left(1/\sqrt{\n}+1/\sqrt{\pp}\right)^{1/3}$, $\widetilde{\kappa}=79.6595$, $\widetilde{\theta}= 0.101037$ and $\widetilde{\alpha}=9.81961$. Thus the first three moments of $\lone$ can be approximated by
\begin{align}
\mlonek{1} \approx &\, \lambda_{np} + \sigma_{np}\, \mgammak{1}\\
\mlonek{2} \approx &\, \lambda_{np}^2 + 2 \lambda_{np}\, \sigma_{np}\, \mgammak{1} + \sigma_{np}^2\, \mgammak{2}\\
\mlonek{3}  \approx &\, \lambda_{np}^3 + 3 \lambda_{np}^2 \, \sigma_{np} \,\mgammak{1} \nonumber\\
& + 3 \lambda_{np} \,\sigma_{np}^2 \,\mgammak{2}+ \sigma_{np}^3\, \mgammak{3}
\end{align}
where $\lambda_{np}=\mu_{np}-\widetilde{\alpha}\,\sigma_{np}$, and the moments of a gamma distributed \ac{r.v.} are given by $\mgammak{i}= {\widetilde{\theta}}^i \,\Gammaf{\widetilde{\kappa}+i} / \Gammaf{\widetilde{\kappa}}$, $\forall i \in \mathbb{N}$.

Alternatively, when $\n$ and $\pp$ are not large, the moments can be computed using numerical integration as
\begin{align} \label{eq:mom_lone_int}
\mlonek{i} = \int_{0}^{\infty} \left(1 - F_{\lone}\left(x^{1/i}\right)\right) dx
\end{align}
using the efficient computation of the \ac{CDF} of $\lone$ proposed in \cite{Chi:14}.

\section*{Appendix B}
\label{sec:appendixB}

Let us denote with $\mathcal{S}_{k}$ the row space of $\Hk{k}$ and with $\mathcal{S}_{k}^{\perp}$ the corresponding orthogonal space. 
Given the assumption of nested models, $\Hk{j}$ is a submatrix of $\Hk{k}$ with $k>j$, and thus
$\mathcal{S}_{j}\subset\mathcal{S}_{k}$ and $\mathcal{S}_{k}^{\perp}\subset\mathcal{S}_{j}^{\perp}$.
Considering \eqref{eq:projmatrix}, we can see that $\Mproj{k}$ is the projection matrix on $\mathcal{S}_{k}^{\perp}$, and thus it is idempotent and symmetric with rank $\n- 2 k$. Also note that $\svv\in\mathcal{S}_{q}$. We then have the following original theorem. 
 
\theorem{
\label{theo:1}
Consider $\MMk{0}$ and $\MMk{1}$, projection matrices on the spaces $\Omega_{0}$ and $\Omega_{1}$, respectively, with $\Omega_{1}\subset\Omega_{0}\subset\mathbb{C}^{\n}$. Given the random row vector $\yvv\sim\CN{\muv}{\sig \Id{\n}}$, with $\muv\in\Omega_{0}^{\perp}$, the \ac{r.v.}
\begin{align} \label{eq:R}
R=\frac{\yvv\MMk{1}\hmt{\yvv}}{\yvv\MMk{0}\hmt{\yvv}}
\end{align}
is beta distributed with parameters $r_{1}$ and $r_{0}-r_{1}$, where $r_{0}$ and $r_{1}$ are the ranks of $\MMk{0}$ and $\MMk{1}$, respectively.
} 

\begin{IEEEproof} 
Let us rewrite $\yvv\MMk{0}\hmt{\yvv}$ as $\yvv\MMk{1}\hmt{\yvv}+\yvv(\MMk{0}-\MMk{1})\hmt{\yvv}$, where $\MMk{0}-\MMk{1}$ is the projection matrix on the orthogonal complement of $\Omega_{1}$ to $\Omega_{0}$. Given \cite[Theorem 4.4.2]{Gra:76} the quadratic forms $\yvv\MMk{0}\hmt{\yvv}$ and $\yvv\MMk{1}\hmt{\yvv}$ are chi squared distributed \acp{r.v.} with $2 r_{0}$ and $2 r_{1}$ \acl{d.o.f.}, respectively, and, given the assumptions, non centrality parameter $\muv\MMk{0}\hmt{\muv}=\muv\MMk{1}\hmt{\muv}=0$. 
Similarly, we can see that $\yvv(\MMk{0}-\MMk{1})\hmt{\yvv}\sim\chisquared{2(r_{0}-r_{1})}$. Due to the properties of projection matrices and the fact that $\Omega_{1}\subset\Omega_{0}$, we have $\MMk{1}(\MMk{0}-\MMk{1})=\MMk{1}\MMk{0}-\MMk{1}=\MMk{1}-\MMk{1}=0$, and thus the quadratic forms $\yvv\MMk{0}\hmt{\yvv}$ and $\yvv\MMk{1}\hmt{\yvv}$ are independent thanks to \cite[Theorem  4.5.3]{Gra:76}. Then the ratio in \eqref{eq:R} can be rewritten as a combination of independent chi squared \acp{r.v.} as
\begin{align} \label{eq:Rbeta}
R=\frac{\yvv\MMk{1}\hmt{\yvv}}{\yvv\MMk{1}\hmt{\yvv}+\yvv(\MMk{0}-\MMk{1})\hmt{\yvv}}
\end{align}
and thus $R\sim\betadist{r_{1}}{r_{0}-r_{1}}$ \cite[Section 26.5]{ZelSev:64}.
\end{IEEEproof}

Thanks to this theorem $R_i$ in \eqref{eq:Ri} is a beta distributed \ac{r.v.} with parameters $\n-2(\q-i)$ and $2i$. It is easy to see that Theorem~\ref{theo:1} can be demonstrated also in the real case, in which $R\sim\betadist{r_{1}/2}{(r_{0}-r_{1})/2}$.

\bibliographystyle{IEEEtran}
\bibliography{IEEEabrv,ITC,bibAM}

\end{document}

%% file: mathdef.tex
\newtheorem{theorem}{Theorem}

\newcommand{\Gammaf}[1]{\Gamma\!\left(#1\right)}                  
\newcommand{\IncLowerGammaf}[2]{\gamma\!\left(#1,#2\right)}              	

\newcommand{\Betaf}[2]{B\!\left(#1,#2\right)}                       			
\newcommand{\BetaIf}[3]{{I}_{#1}\!\left(#2,#3\right)}

\newcommand{\Prob}[1]{\mathbb{P}\!\left(#1\right)}
\newcommand{\CN}[2]{\mathcal{CN}\!\left(#1,#2\right)}       	
\newcommand{\chisquared}[1]{\chi_{#1}^2}                   		
\newcommand{\betadist}[2]{\beta_{#1,#2}}                   		
\newcommand{\EX}[1]{{\mathbb{E}}\left\{{#1}\right\}}        		
\newcommand{\mk}[1]{\mathsf{m}_{#1}}                        		

\newcommand{\fcond}[2]{f\!\left(#1;#2\right)}                   	
\newcommand{\lnfcond}[2]{\ln f\!\left(#1;#2\!\right)}           	

\newcommand{\hmt}[1]{{#1}^{\textrm{H}}}  		
\newcommand{\transp}[1]{{#1}^{\textrm{T}}}               
\newcommand{\tr}[1]{\textrm{tr}\!\left\{#1\right\}}         	

\newcommand{\Id}[1]{{\mathbf I}_{#1}}                                        		
\newcommand{\Cv}[1]{{\mathbb{C}}^{#1}}                                      	

\def\SNR{\textsf{SNR}}

\def\sig{\sigma^2}

\newcommand{\xv}[1]{\mathbf{x}_{#1}}
\newcommand{\nv}[1]{\mathbf{n}_{#1}}
\newcommand{\sv}[1]{\mathbf{s}_{#1}}

\def\xvi{\mathbf{x}_{i}}
\def\nvi{\mathbf{n}_{i}}

\def\Y{\mathbf{Y}}                      

\def\Hm{\mathbf{H}}                     

\def\Sig{\mathbf{\Sigma}}

\def\SCM{\mathbf{S}}                    
\def\R{\mathbf{R}}                      

\def\xiq{x_{i,q}}

\def\svi{{\mathbf{s}}_{i}}
\def\zvi{{\mathbf{z}}_{i}}

\def\Xv{\mathbf{X}}

\def\thetak{{\boldsymbol\Theta}^{(k)}}
\def\thetakest{\widehat{\boldsymbol{\Theta}}^{\left(k\right)}}

\def\PAICk{\mathscr{P}_{\textrm{AIC}}\!\left(k\right)}

\def\PBICk{\mathscr{P}_{\textrm{BIC}}\!\left(k\right)}
\def\PGICk{\mathscr{P}_{\textrm{GIC}}\!\left(k\right)}

\def\ITCk{\textsf{ITC}\!\left(k\right)}
\newcommand{\ITCkk}[1]{\textsf{ITC}\!\left(#1\right)}

\def\phik{\phi\!\left(k\right)}

\def\Pc{\text{P}_{\textrm{c}}}
\def\PcDES{\text{P}_{\text{c}}^\text{DES}}

\def\Pover{\text{P}_{\textrm{over}}}
\def\PoverMAX{\text{P}_{\text{over}}^\text{MAX}}
\def\PoverUB{\text{P}_{\textrm{over}}^{\text{UB}}}
\def\PoverLB{\text{P}_{\textrm{over}}^{\text{LB}}}

\def\Punder{\text{P}_{\textrm{under}}}

\newcommand{\Pkk}[1]{\mathscr{P}\!\left(#1\right)}

\def\xiq{\xi_{\q}}

\def\nutilde{\widetilde{\nu}}

\def\imax{i_{\text{max}}}
\newcommand{\mlonek}[1]{\mathsf{m}_{#1}^{(\lone)}}                        	
\newcommand{\mtk}[1]{\mathsf{m}_{#1}^{(\t)}}                        		
\newcommand{\mgammak}[1]{\mathsf{m}_{#1}^{(\Gamma)}}		

\def\n{{n}}
\def\q{{q}}
\def\qmax{{q}_{\text{max}}}
\def\qest{\widehat{\q}}
\def\qstar{{\q}^{*}}
\def\p{{p}}
\def\pp{{p}'}
\newcommand{\vv}[1]{\mathbf{v}_{#1}}
\def\thetaq{{\boldsymbol\Theta}^{(\q)}}

\def\li{l_{i}}
\def\lone{\ell_{1}}
\newcommand\lkk[1]{l_{#1}}
\def\u{u}

\def\v{v}
\def\vroot{\text{v}}

\def\vtildeqstar{\widetilde{\text{v}}_{\qstar}}

\def\t{t}

\newcommand{\Hk}[1]{\mathbf{H}_{#1}}
\newcommand{\HkH}[1]{\mathbf{H}_{#1}^{\text{H}}}
\newcommand{\hvki}[1]{\mathbf{h}_{#1,(k)}}

\newcommand{\fii}[1]{f_{#1}}
\newcommand{\phaseii}[1]{\varphi_{#1}}
\newcommand{\aii}[1]{a_{#1}}
\def\nvv{\mathbf{n}}
\def\yvv{\mathbf{y}}
\def\svv{\mathbf{s}}

\def\betavv{\boldsymbol{\beta}}
\newcommand{\Mproj}[1]{\mathbf{P}_{#1}^{\perp}}

\newcommand{\MMk}[1]{\mathbf{M}_{#1}}
\def\sigestk{\widehat{\sigma}_{k}^2}

\def\muv{\boldsymbol{\mu}}
\def\psiq{\psi\!\left(q\right)}



%% file: acronymsAM.tex
\begin{acronym}
\scriptsize
\acro{ADC}{analog to digital converter}
\acro{AWGN}{additive white Gaussian noise}
\acro{CDF}{cumulative distribution function}
\acro{c.d.f.}{cumulative distribution function}
\acro{CR}{cognitive radio}
\acro{DAA}{detect and avoid}
\acro{DFT}{discrete Fourier transform}
\acro{DVB-T}{digital video broadcasting\,--\,terrestrial}
\acro{ECC}{European Community Commission}
\acro{FCC}{Federal Communications Commission}
\acro{FFT}{fast Fourier transform}
\acro{GPS}{Global Positioning System}
\acro{i.i.d.}{independent, identically distributed}
\acro{IFFT}{inverse fast Fourier transform}
\acro{ITU}{International Telecommunication Union}
\acro{MB}{multiband}
\acro{MC}{multicarrier}
\acro{MF}{matched filter}
\acro{MI}{mutual information}
\acro{MIMO}{multiple-input multiple-output}
\acro{MRC}{maximal ratio combining}
\acro{MMSE}{minimum mean-square error}
\acro{OFDM}{orthogonal frequency-division multiplexing}
\acro{p.d.f.}{probability density function}
\acro{PAM}{pulse amplitude modulation}
\acro{PSD}{power spectral density}
\acro{PSK}{phase shift keying}
\acro{QAM}{quadrature amplitude modulation}
\acro{QPSK}{quadrature phase shift keying}
\acro{r.v.}{random variable}
\acro{R.V.}{random vector}
\acro{SNR}{signal-to-noise ratio}
\acro{SS}{spread spectrum}
\acro{TH}{time-hopping}
\acro{ToA}{time-of-arrival}
\acro{UWB}{ultrawide band}
\acro{UWBcap}[UWB]{Ultrawide band}
\acro{WLAN}{wireless local area network}
\acro{WMAN}{wireless metropolitan area network}
\acro{WPAN}{wireless personal area network}
\acro{WSN}{wireless sensor network}

\acro{ML}{maximum likelihood}
\acro{GLR}{generalized likelihood ratio}
\acro{GLRT}{generalized likelihood ratio test}
\acro{LLRT}{log-likelihood ratio test}
\acro{LRT}{likelihood ratio test}
\acro{$P_{EM}$}{probability of emulation, or false alarm}
\acro{$P_{MD}$}{probability of missed detection}
\acro{$P_D$}{probability of detection}
\acro{$P_{FA}$}{probability of false alarm}
\acro{ROC}{receiver operating characteristic}

\acro{AGM}{arithmetic-geometric mean ratio test}
\acro{AIC}{Akaike information criterion}
\acro{BIC}{Bayesian information criterion}
\acro{BWB}{bounded worse beaviour}
\acro{CA-CFAR}{cell-averaging constant false alarm rate}
\acro{CAIC}{consistent AIC}
\acro{CAICF}{consistent AIC with Fisher information}
\acro{CFAR}{constant false alarm rate}
\acro{CDR}{constant detection rate}
\acro{CP}{cyclic prefix}
\acro{CPC}{cognitive pilot channel}
\acro{CL}{centroid localization}
\acro{CSCG}{circularly-symmetric complex Gaussian}
\acro{d.o.f.}{degrees of freedom}
\acro{DC}{distance based combining}
\acro{DSM}{dynamic spectrum management}
\acro{DSA}{dynamic spectrum access}
\acro{DTV}{digital television}
\acro{ECC}{European Communications Committee}
\acro{ED}{energy detector}
\acro{EDC}{efficient detection criterion}
\acro{EGC}{equal gain combining}
\acro{EME}{energy to minimum eigenvalue ratio}
\acro{ENP}{estimated noise power}
\acro{ERC}{European Research Council}
\acro{ES-WCL}{energy based selection WCL}
\acro{EU}{European Union}
\acro{ExAIC}{exact AIC}
\acro{ExBIC}{exact BIC}
\acro{FC}{fusion center}
\acro{FIM}{Fisher information matrix}
\acro{GLM}{general linear model}
\acro{GLRT}{generalized likelihood ratio test}
\acro{GIC}{generalized information criterion}
\acro{HC}{hard combining}
\acro{HF}{hard fusion}
\acro{ITC}{information theoretic criteria}
\acro{K-L}{Kullback-Leibler}
\acro{LR}{likelihood ratio}
\acro{LS}{least square}
\acro{MDL}{minimum description length}
\acro{MET}{ratio of maximum eigenvalue to the trace}
\acro{ML}{maximum likelihood}
\acro{MTM}{multi taper method}
\acro{MME}{maximum to minimum eigenvalues ratio}
\acro{NP}{Neyman-Pearson}
\acro{OFCOM}{Office of Communications}
\acro{OSA}{opportunistic spectrum access}
\acro{OSF}{oversampling factor}
\acro{PMSE}{programme making and special events}
\acro{PU}{primary user}
\acro{PWCL}{power based WCL}
\acro{r.o.f.}{roll-off factor}
\acro{RAN}{Regional Area Network}
\acro{RMSE}{root mean square error}
\acro{RWCL}{relative WCL}
\acro{RSS}{received signal strength}
\acro{SC}{soft combining}
\acro{SCM}{sample covariance matrix}
\acro{SCovM}{sample covariance matrix}
\acro{SCorM}{sample correlation matrix}
\acro{SDR}{software defined radio}
\acro{SF}{soft fusion}
\acro{SP}{sensing period}
\acro{SS}{spectrum sensing}
\acro{SSE}{signal subspace eigenvalues}
\acro{SU}{secondary user}
\acro{TOA}{time of arrival}
\acro{TDOA}{time difference of arrival}
\acro{TVBD}{TV Bands Device}
\acro{TS-WCL}{two step WCL}
\acro{WCL}{weighted centroid localization}
\acro{WSD}{white space device}
\acro{WSN}{wireless sensor network}
\acro{WSS}{wideband spectrum sensing}

\end{acronym}